\documentclass{article}

\usepackage{microtype}
\usepackage{graphicx}
\usepackage{subfigure}
\usepackage{booktabs} 

\usepackage{hyperref}

\usepackage[accepted]{icml2025}

\usepackage{amsmath}
\usepackage{amssymb}
\usepackage{mathtools}
\usepackage{amsthm}
\usepackage{enumitem}
\usepackage{pifont}
\usepackage[capitalize,noabbrev]{cleveref}

\usepackage{smile}
\usepackage{tabularray} 
\UseTblrLibrary{booktabs}
\definecolor{LightCyan}{rgb}{0.8, 0.9, 1}
\definecolor{maroon}{cmyk}{0,0.87,0.68,0.32}
\newcolumntype{a}{>{\columncolor{LightCyan}}c}
\newcommand{\XSolidBrush}{\ding{55}}

\usepackage[textsize=tiny]{todonotes}
\newcommand{\method}{\texttt{MARINE}}
\usepackage[compact]{titlesec}
\usepackage{sidecap}
\titlespacing*{\section}{5pt}{*0.3}{*0.3}
\titlespacing*{\subsection}{4pt}{*0.3}{*0.3}

\icmltitlerunning{Mitigating Object Hallucination in Large Vision-Language Models via Image-Grounded Guidance}

\begin{document}

\twocolumn[
\icmltitle{Mitigating Object Hallucination in Large Vision-Language Models via Image-Grounded Guidance}



\icmlsetsymbol{equal}{*}

\begin{icmlauthorlist}
\icmlauthor{Linxi Zhao}{equal,cornell}
\icmlauthor{Yihe Deng}{equal,ucla}
\icmlauthor{Weitong Zhang}{unc}
\icmlauthor{Quanquan Gu}{ucla}
\end{icmlauthorlist}

\icmlaffiliation{cornell}{Department of Computer Science, Cornell University, Ithaca, NY, USA}
\icmlaffiliation{ucla}{Department of Computer Science, University of California, Los Angeles, CA, USA}
\icmlaffiliation{unc}{School of Data Science and Society, UNC, Chapel Hill, NC, USA}
\icmlcorrespondingauthor{Quanquan Gu}{qgu@cs.ucla.edu}

\icmlkeywords{Machine Learning, ICML}

\vskip 0.3in
]



\printAffiliationsAndNotice{\icmlEqualContribution}

\begin{abstract}
The advancement of Large Vision-Language Models (LVLMs) has increasingly highlighted the critical issue of their tendency to hallucinate non-existing objects in the images. 
To address this issue, previous works focused on using specially curated datasets or powerful LLMs to rectify the outputs of LVLMs. 
However, these approaches require either costly training or fine-tuning, or API access to proprietary LLMs for post-generation correction.
In response to these limitations, we propose \textbf{M}itigating hallucin\textbf{A}tion via image-g\textbf{R}ounded gu\textbf{I}da\textbf{N}c\textbf{E} ($\method$), a framework that is both \emph{training-free} and \emph{API-free}. 
$\method$ effectively and efficiently reduces object hallucinations during inference by introducing image-grounded guidance to LVLMs. 
This is achieved by leveraging open-source vision models to extract object-level information, thereby enhancing the precision of LVLM-generated content.
Our framework's flexibility further allows for the integration of multiple vision models, enabling more reliable and robust object-level guidance.  
Through comprehensive evaluations across $5$ popular LVLMs with diverse evaluation metrics and benchmarks, we demonstrate the effectiveness of $\method$, which even outperforms existing fine-tuning-based methods. 
Remarkably, it reduces hallucinations consistently in GPT-4V-assisted evaluation while maintaining the detailedness of LVLMs' generations. 
We release our code at \url{https://github.com/Linxi-ZHAO/MARINE}.\end{abstract}

\section{Introduction}
The advent of Large Language Models (LLMs) has motivated advancements in extending their remarkable capabilities to multimodal data. 
Grounded in the development of pre-trained vision-language models~\citep{radford2021learning,jia2021scaling,alayrac2022flamingo} that align visual and textual embedding spaces, Large Vision Language Models (LVLMs) have gained substantial attention in both architectural development~\citep{liu2023llava,zhu2023minigpt,ye2023mplug,dai2023instructblip,gao2023llamaadapter}, alignment~\citep{yu2024rlhf,zhou2024aligning,deng2024enhancing} and benchmarking datasets~\citep{xu2023lvlm,lu2024mathvista,zhang2024mathverse}. 
However, similar to the hallucination issues in textual LLMs ~\citep{ji2023survey}, where irrelevant content is generated with input prompts, LVLMs face a specific challenge known as object hallucination: generating non-existing objects for a given image~\citep{li2023evaluating,wang2023evaluation, zhou2023analyzing,fu2023mme,lovenia2023negative, jing2023faithscore}. 
Such a problem is particularly concerning as it compromises the model's accuracy and reliability, especially considering the growing application of LVLMs to safety-critical downstream tasks such as medical imaging~\citep{chambon2022adapting,bazi2023vision}.

\begin{figure*}[!ht]
    \centering
    \includegraphics[width=\textwidth]{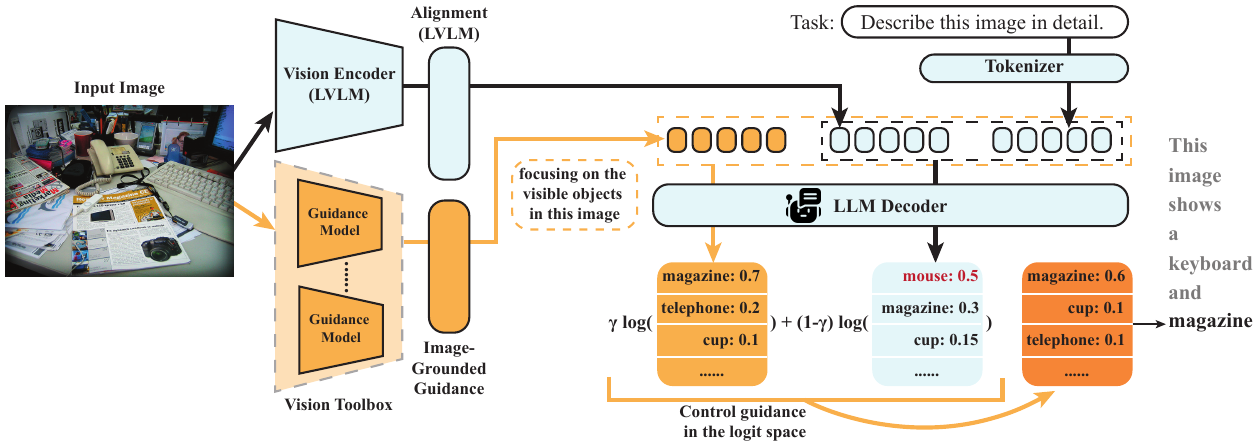}
    \caption{Illustration of $\method$ framework, which introduces a vision toolbox with one or multiple guidance models to enrich the visual context of the original LVLM. The output logits are controlled to place more importance on the guided generation with the guidance strength $\gamma$.}
    \label{fig:demo}
\end{figure*}

In response to the pressing issue of object hallucinations in LVLMs, early attempts~\citep{liu2023aligning,liu2023mitigating,gunjal2023detecting,wang2023vigc} focused on addressing the bias by curating high-quality datasets for fine-tuning or leveraging advanced GPT queries~\citep{yin2023woodpecker}, such as GPT-4, to post-process the generated captions. However, these methods can be infeasible to implement. For instance, creating extensive, high-quality datasets for fine-tuning LVLMs is costly and requires significant human annotation. Additionally, relying on advanced GPT models for post-processing is expensive and can raise privacy concerns, especially in sensitive fields like medical imaging. Most importantly, these approaches do not address the \emph{intrinsic} causes of object hallucination in LVLMs. 

In this paper, we investigate the intrinsic causes of object hallucination in LVLMs. Specifically, these deficiencies may stem from the three main components of the LVLMs: 1) insufficient visual context provided by the visual encoder~\citep{zhang2023multimodal}, 2) distortion or loss of visual information during the projection from vision to text space, and 3) inherent hallucinations common in general language models. 
To address the first two LVLM-specific causes, we introduce \textbf{M}itigating hallucin\textbf{A}tion via image-g\textbf{R}ounded gu\textbf{I}da\textbf{N}c\textbf{E} ($\method$). \method~mitigates hallucination issues arising from the visual encoder and information distortion during cross-modal alignment by leveraging external guidance from image-grounded models, such as object detection models. 
Our approach leverages the inherent advantage of these image-grounded models, which are specifically designed and trained for more detailed visual information extraction. These models provide higher quality, fine-grained visual encoding compared to the standard visual encoders in LVLMs, which are primarily optimized for grasping the overall context of an image. 
Furthermore, we integrate the guidance from image-grounded models into text descriptions, allowing the LVLM to process the information without requiring additional alignment procedures. 
As a result, \method~is a training-free, API-free
method that addresses object hallucination at inference time by targeting its two root causes. 

As shown in Figure~\ref{fig:demo}, \method~incorporates one or more image-grounding models to enrich the visual context of LVLMs. The guidance are then aggregated as prompt input to the LLM decoder to improve the response quality. Empirical evaluations are conducted on five widely-recognized LVLMs across benchmarks 
including MSCOCO~\citep{lin2014microsoft}, LLaVA-QA90 task~\citep{liu2023llava}, A-OKVQA~\citep{schwenk2022okvqa}, and GQA~\citep{hudson2019gqa}. 
We present results based on guidance from a aggregated source of DEtection TRansformer (DETR)~\citep{carion2020end} and RAM++~\citep{huang2023opensetimagetaggingmultigrained}. We also include ideal results based on ground truth object oracle, denoted as $\method$-Truth. 
Our experimental results demonstrate that, in comparison with state-of-the-art algorithms, \method~exhibits further reduced hallucination, as measured by popular hallucination metrics 
such as CHAIR~\citep{rohrbach2018object} and POPE~\citep{li2023evaluating}, as well as GPT-4V's evaluation.  
These results confirm that \method~can effectively mitigate object hallucinations without requiring additional training resources or access to proprietary LLMs. 
To summarize, our contribution are listed as follows:
\begin{itemize}[leftmargin=*,nosep]
\item We introduce $\method$, a universal framework and aggregating a toolbox of image-grounded visual models to guide the generation process of LVLMs. 
$\method$ leverages the intrinsic advantages of these visual models in providing the detailed information of the input image and help mitigate the hallucinations in LVLMs. 
\item Through extensive evaluations on various datasets, we demonstrate that $\method$ consistently outperform the baselines in hallucination mitigation while maintaining overall performance across multiple tasks (image captioning, VQA). 
\item $\method$ provides a favorable trade-off between latency and accuracy, with the lowest computational overhead compared to existing baselines, which 
positions $\method$ as a practical and scalable solution for real-world applications without significant computational cost.
\end{itemize}

\section{Related Work}
\subsection{Object Hallucination in Large Vision-Language Models}
The hallucination issue in Large Vision-Language Models (LVLMs) \citep{liu2023llava,zhu2023minigpt,ye2023mplug,dai2023instructblip,gao2023llamaadapter} has drawn significant attention, as highlighted by studies~\citep{li2023evaluating, wang2023evaluation, zhou2023analyzing, fu2023mme, lovenia2023negative}.
Notably, different from textual LLMs, LVLMs are prone to a unique type of hallucination called `object hallucination' \citep{rohrbach2018object}, where the model falsely perceives the presence of non-existent objects in images. 
Efforts to address this problem in LVLMs include fine-tuning approaches using vision-language datasets \citep{liu2023mitigating, gunjal2023detecting}, as well as GPT-assisted methods such as those by \citet{zhai2023halle}. Notably, \citet{yin2023woodpecker} proposed a training-free approach using GPT-3.5 for hallucination correction.

Concurrently, \citet{leng2023mitigating} introduced Visual Contrastive Decoding (VCD), a technique that applies noise to image inputs and penalizes logit outputs of these corrupted images. \citet{huang2023opera} enhanced beam-search decoding with the Over-trust Penalty and Retrospection-Allocation Strategy (OPERA), which penalizes over-trust and refines token selection based on previous outputs. HALC \citep{chen2024halc} employs adaptive focal-contrast decoding to encourage LVLMs to focus on fine-grained visual information, while using a computationally intensive beam search algorithm. 
In addition, BRAVE~\citep{kar2024bravebroadeningvisualencoding} introduces a new architecture that combines features from multiple vision encoders. While not directly targeting hallucination, it shares the key insight of leveraging diverse visual signals to improve grounding.

\subsection{Controllable Generation} 
Controllable text generation~\citep{prabhumoye2020exploring,hu2021causal,zhang2023survey} has emerged as a vital research domain, focusing on the generation of natural sentences with controllable attributes such as persona~\citep{prabhumoye2020exploring,hu2021causal,zhang2023survey} and politeness~\citep{niu2018polite,madaan2020politeness}. 
Among the various approaches, fine-tuning has been recognized as the most straightforward approach, achieved either through full fine-tuning~\citep{li2021prefix,ouyang2022training,carlsson2022fine} or integrating tunable adaptors~\citep{lin2021adapter,ribeiro2021structural}.
While fine-tuning has been effective in a wide range of applications, it is also expensive in computation as the size of LLMs is growing tremendously. 
Recently, there has been a development on controllable generation with diffusion models~\citep{li2022diffusion,lin2023text}, extending to controllable text-to-image generation~\citep{yang2023reco}. 
Particularly, the use of classifier guidance~\citep{dhariwal2021diffusion} and classifier-free guidance~\citep{ho2021classifier} has become prominent in refining the quality of generated outputs. 
Most recently, \citet{sanchez2023stay} applied classifier-free guidance to language models in the \emph{single-modal} setting to improve their performance at inference time. 
Our approach methodologically resembles classifier-free guidance for LVLMs' text generation, while specifically addressing the \emph{multi-modal} context and focusing on reducing hallucinations.

\section{Preliminaries}

\paragraph{Generative language models.}
Let $p_{\btheta}$ denotes an LLM parameterized by $\btheta$. Consider a sequence $\xb=[x_1,\ldots,x_n]$ as the input prompt, where each $x_i$ is a token from a predefined vocabulary.  
The LLM then generates the response sequence $\yb=[y_1,\ldots,y_m]$ by sampling from the conditional probability distribution $p_{\btheta}(\cdot|\xb)$, where $y_t$ denotes individual token for $1\le t\le m$. 
The conditional distribution $p_{\btheta}(\yb|\xb)$ can therefore be expressed as $p_{\btheta}(\yb|\xb) = \prod_{t=1}^{m}p_{\btheta}(y_{t}|\xb, \yb_{<t})$,
where $\yb_{<t}=[y_1,\ldots,y_{t-1}]$ for $t>1$ and is empty for $t=1$. In the case of LVLMs, visual tokens $\vb=[v_1,\ldots,v_k]$ are additionally included. These tokens are generated from a pre-trained visual encoder and mapped into the token space through a linear projection. The conditional distribution of output $\yb$ given the visual tokens $\vb$ and textual prompt $\xb$ is expressed as $\textstyle{p_{\btheta}(\yb|\vb,\xb) = \prod_{t=1}^{m}p_{\btheta}(y_{t}|\vb,\xb, \yb_{<t}),}$ where $p_{\btheta}$ is approximated by LVLMs. 

\paragraph{Guidance in generative models.} 
The process of a guided generation involves getting the output $\yb$ conditioned on input $\xb$, which encodes the desired properties of the output $\yb$. This guidance can be generally added to the model by two distinct approaches: classifier guidance~\citep{dhariwal2021diffusion} and classifier-free guidance~\citep{ho2021classifier}. As a top-level view, both methods formulate the conditional probability distribution of output $\yb$ conditioned on guidance $\xb$ as
\begin{align}
    p(\yb | \xb) \propto p_{\btheta}(\yb) p(\xb | \yb)^\gamma,\label{eq:guidance}
\end{align}
where $p_{\btheta}(\yb)$ is the original generative model and $p(\xb | \yb)$ is the posterior distribution of $\xb$ given $\yb$ and $\gamma$ is the guidance strength. In the classifier guidance, the posterior distribution $p(\xb | \yb)$ in~\eqref{eq:guidance} is replaced by a classifier $p_{\bphi}(\xb | \yb)$ parameterized by $\bphi$, which requires additional training step and calculating $\nabla_{\xb} \log p_{\bphi}(\xb | \yb)$. The classifier-free guidance, on the other hand, removes the necessity of the parameterized classifier $f_{\bphi}$. Instead, according to the Bayes rule, the posterior distribution can be approximated by $p_{\btheta}(\xb | \yb) \propto {p_{\btheta}(\yb | \xb)} / {p_{\btheta}(\yb)}$,
where $p_{\btheta}(\yb | \xb)$ is the generative model when taking $\xb$ as prompt input. Plugging this back into~\eqref{eq:guidance} yields the guided distribution that can be approximated by 
\begin{align*}
    \hat{p}_{\btheta}(\yb|\xb) \propto p_{\btheta}(\yb) \cdot \frac{p_{\btheta}(\yb|\xb)^{\gamma}}{p_{\btheta}(\yb)^{\gamma}} 
    = \frac{p_{\btheta}(\yb|\xb)^{\gamma}}{p_{\btheta}(\yb)^{\gamma-1}}.
\end{align*}

As a result, the guided LLM $\hat{p}_{\btheta}$ places more importance on the prompt $\xb$ during generation with the increasing value of $\gamma$, thereby producing texts that better align with the desired behavior from the prompt~\citep{sanchez2023stay}.

\section{Method}\label{sec:meth}
The existing architecture of LVLMs is composed of a visual encoder, a visual and textual domain alignment layer, and the LLM itself. Therefore, besides the inherent language priors of LLMs~\citep{biten2022let}, object hallucination may arise from (1) deficiencies in the visual encoder provide insufficient visual information~\citep{zhang2023multimodal} and (2) distortion or loss of visual information during the projection from vision to language space. 
To mitigate object hallucinations, we introduce \method, a framework containing two major components to address the previous challenges: (1) introducing additional visual information from a set of vision models and (2) using the additional aggregated visual features to guide the LVLM's generation. In Figure~\ref{fig:demo}, we present the framework overview. 

\subsection{Visual Guidance from Image-Grounded Features}
To introduce image-grounded guidance to mitigate hallucinations, our approach integrates additional object detection models, which differ from the visual encoders used in LVLM that are usually pre-trained from CLIP~\citep{radford2021learning}. 
This integration leverages object detection models to extract detailed visual information from images. Upon acquiring extra visual information from different image-grounded models, we aggregate and translate the collected information into textual information. This aggregation can be done by the language model~\citep{lin2023aggretriever} or rule-based algorithm~\citep{bird2009natural}. Such an information aggregation is effective and efficient, as it eliminates the necessity of fine-tuning the alignment layer while retaining the rich information encoded by various of image grounding models. 
We subsequently employ a simple prompt ``focusing on the visible objects in this image:'' and concatenate it with the aggregated object information, denoted as the guidance prompt $\cbb$.

\subsection{Guided Text Generation with Visual Information}
We tackle the object hallucination problem of LVLMs by placing importance on additional image-grounded information.  
In addition to the visual tokens $\vb$ extracted from the original LVLM and textual prompt $\xb$, we extract the auxiliary visual tokens $\cbb$ from the additional guidance models. 
The generation of the $t$-th token in the output $\yb$ of our classifier-free guided LVLM $p_{\btheta}$ is expressed as 
\begin{align*}
    \hat{p}_{\btheta}(y_t|\vb,\cbb,\xb,\yb_{<t}) \propto \frac{p_{\btheta}(y_t|\vb,\cbb,\xb,\yb_{<t})^{\gamma}}{p_{\btheta}(y_t|\vb,\xb,\yb_{<t})^{\gamma-1}},
\end{align*}
where $\cbb$ denotes our control guidance and $\gamma$ is the control strength. The sampling of output generation is given by 
\begin{align*}
    \hat{p}_{\btheta}(\yb|\vb,\cbb,\xb) &= \textstyle{\prod_{t=1}^m} \hat{p}_{\btheta}(y_t|\vb,\cbb,\xb,\yb_{<t}) \\
    &\propto \textstyle{\prod_{t=1}^m} \frac{p_{\btheta}(y_t|\vb,\cbb,\xb,\yb_{<t})^{\gamma}}{p_{\btheta}(y_{t}|\vb,\xb,\yb_{<t})^{\gamma-1}} \\
    &= \frac{p_{\btheta}(\yb|\vb,\cbb,\xb)^\gamma}{p_{\btheta}(\yb|\vb,\xb)^{\gamma-1}}.
\end{align*}

We can further view $\method$ in the logit space, where the $t$-th token is therefore sampled from the logit space by
\begin{align*}
    \log\hat{p}_{\btheta}(y_t|\vb,\cbb,\xb,\yb_{<t}) &= \gamma\log p_{\btheta}(\yb|\vb,\cbb,\xb,\yb_{<t}) \\
    &\quad+(1-\gamma)\log p_{\btheta}(\yb|\vb,\xb,\yb_{<t}).
\end{align*}
This linear combination of logits implies that the conditional generation on the additional image-grounded guidance acts as a controllable gate. 
Only objects with relatively high probabilities in both branches could appear at top when sampling.
Specifically, setting $\gamma=0$ recovers the original LLM generation without control guidance and setting $\gamma=1$ produces the LLM generation entirely based on the control.  
Meanwhile, for $\gamma \in (0,1)$, $\method$ yields a combination of the original generation $p_{\btheta}(\yb|\vb,\xb)$ and the generation conditioned on the guidance $p_{\btheta}(\yb|\vb,\cbb,\xb)$. 
This strikes a balance between a better ability to follow instructions to generate high-quality answers and the increased accuracy and detail in image descriptions. 
The formulation therefore shares resemblance to the classifier-free guidance introduced for LLMs~\citep{sanchez2023stay}, which places importance on the textual prompt itself to better align the LLM generation with user intention in the \emph{single-modal} setting. 
We summarize $\method$ in Algorithm~\ref{alg:method}. In detail, \method~aggregates the collected visual information $\{\cbb_i\}_i$ using function $\mathrm{Aggr.}$, which can be a small language model for information aggregation~\citep{lin2023aggretriever}.

\begin{algorithm}[!hb]
\caption{\textbf{M}itigating hallucin\textbf{A}tion via image-g\textbf{R}ounded gu\textbf{I}da\textbf{N}c\textbf{E} ($\method$)}\label{alg:method}
\begin{algorithmic}[1]
\STATE \textbf{Input:} LLM parameter $\btheta$, input prompt $\xb$, visual tokens $\vb$ from LVLM's original vision tower
\STATE \textbf{Input:} auxiliary visual tokens $\{\cbb_i\}_{i=1}^M$ from $M$ image grounding models, guidance scale $\gamma$
\STATE Initialize empty output $\yb=[]$. 
\STATE Aggregate visual information as textual prompt $\cbb = \mathrm{Aggr.}(\{\cbb_i\}_{i=1}^M)$ \label{ln:aggr}
\FOR{$t = 0,1,\ldots,T$}
\STATE Construct unconditional input $\xb_{\text{uncond}}^{(t)} = [\vb$, $\xb$, $\yb_{<t}]$. 
\STATE Generate unconditional output logits using LLM:  $\ell_{\text{uncond}}^{(t)}=\log p_{\btheta} (\xb_{\text{uncond}}^{(t)})$. \label{ln:7}
\STATE Construct conditional input $\xb_{\text{cond}}^{(t)} = [\vb$, $\cbb$, $\xb$, $\yb_{<t}]$.
\STATE Generate conditional output logits using LLM: $\ell_{\text{cond}}^{(t)}$ = $\log p_{\btheta} (\xb_{\text{cond}}^{(t)})$. \label{ln:9}
\STATE Update output logits $\ell^{(t)}$ = $\gamma \ell_{\text{cond}}^{(t)} + (1-\gamma)\ell_{\text{uncond}}^{(t)}$.
\STATE Sample token $y_t$ from logit space denoted by $\ell^{(t)}$.
\STATE Let $\yb = [\yb,y_{t}]$.
\ENDFOR
\STATE \textbf{Output:} $\yb$.
\end{algorithmic}
\end{algorithm}

\section{Experiments}
In this section, we evaluate $\method$ in mitigating object hallucinations across various LVLMs, showing that it outperforms state-of-the-art methods on established metrics across different question formats.
\subsection{Experiment Setup}
\noindent\textbf{Models.} 
To demonstrate the broad applicability of our approach across different LVLM architectures, we apply and evaluate $\method$ to widely-used models including \emph{LLaVA}~\citep{liu2023llava}, \emph{LLaVA-v1.5}~\citep{liu2023improved}, \emph{MiniGPT-v2}~\citep{chen2023minigpt}, \emph{mPLUG-Owl2}~\citep{ye2023mplug} and \emph{InstructBLIP}~\citep{liu2023improved}. 
To address the object hallucination problems in text generation, we incorporate the DEtection TRansformer (DETR) \citep{carion2020end} and RAM++~\citep{huang2023opensetimagetaggingmultigrained} as the additional vision models for guidance. 

\paragraph{Guidance from multiple sources.} Our framework's compatibility with various vision models allows for the incorporation of multiple sources to enhance precision and robustness. 
By considering object-level information from DETR and RAM++ simultaneously, we generate guidance that reflects consensus across these models. 
This approach significantly improves the accuracy and reliability of the guidance provided to the LVLM. 

\paragraph{Datasets and evaluations.}
In alignment with established evaluations from previous studies~\citep{dai2023plausible,yin2023woodpecker}, we assess our method using the following metrics:
\begin{itemize}[leftmargin=*,nosep]
    \item Caption Hallucination Assessment with Image Relevance (\emph{CHAIR})~\citep{rohrbach2018object}. 
    It involves prompting the LVLMs to generate a description for the input image, and then comparing this generation with ground truth objects present in the image.
    CHAIR quantifies hallucination both at instance level and sentence level, respectively defined as CHAIR$_I$ and CHAIR$_S$: 
    \begin{gather*}
        \text{CHAIR}_I = \frac{\big|\{\text{hallucinated objects}\}\big|}{\big|\{\text{all mentioned objects}\}\big|}\\
        \text{CHAIR}_S = \frac{\big|\{\text{captions with hallucinated objects}\}\big|}{\big|\{\text{all captions}\}\big|}
    \end{gather*}
    In addition to these metrics, we incorporate an instance-level Recall score in our evaluation to evaluate whether the descriptions accurately include the necessary visual content from the image: 
    \begin{align*}
        \text{Recall} &= \frac{\big|\{\text{non-hallucinated objects}\}\big|}{{\big|\{\text{all existing objects}\}\big|}}
    \end{align*}
    \item Polling-based Object Probing Evaluation (\emph{POPE})~\citep{li2023evaluating}. 
    POPE formulates a binary classification task by prompting LVLMs with questions such as ``Is there a keyboard in this image?'' to answer ``yes'' or ``no''. 
    We specifically focus on the adversarial setting, which is considered the most challenging setting. Results for the random and popular settings are detailed in Appendix~\ref{app:exp_results}.
    We report the accuracy and F1 score of the LVLMs' responses, and the proportion of ``yes'' answers.

    \item \emph{GPT-4V-aided Evaluation}~\citep{yin2023woodpecker}.
    The GPT-4V-aided evaluation compares the outputs of two LVLM assistants using GPT-4V as a judge. 
    In this evaluation, we utilize the LLaVA-QA90 task~\citep{liu2023llava}
    (including conversations, visual perceptions, and complex reasoning tasks) and additionally consider the image captioning task.
\end{itemize}
Consistent with ~\citet{li2023evaluating}, we randomly sampled a subset of 500 images from MSCOCO~\citep{lin2014microsoft} dataset for CHAIR evaluation. For the POPE evaluation, we created 3000 questions across three datasets—500 images each from MSCOCO, A-OKVQA~\citep{schwenk2022okvqa}, and GQA~\citep{hudson2019gqa}. For the GPT-4V-aided evaluation, we utilized 90 questions from the LLaVA-QA90 task and randomly selected 50 MSCOCO images for image captioning task.

\paragraph{Baselines.}
In addition to comparing with the performance of the original LVLM sampling method, we also consider the following popular methods for mitigating hallucinations.  
\begin{itemize}[leftmargin=*,nosep]
    \item \emph{Greedy-Decoding}, which adopts the greedy sampling strategy, by generating tokens with the highest posterior probability to address hallucinations arising from.
    \item \emph{LURE} \citep{zhou2023analyzing}, which identifies and masks potentially hallucinated words and fine-tune a MiniGPT4 model to rectify object hallucinations in the generated descriptions.    
    \item \emph{Woodpecker} \citep{yin2023woodpecker}, which leverages GPT-3.5 to correct hallucinations in LVLM generation with five steps toward the correction.
    \item \emph{VCD}~\citep{leng2023mitigating}, which distorts the image inputs to impose penalties on logit outputs. 
    \item \emph{OPERA}~\citep{huang2023opera}, which penalizes logits to mitigate over-trust in beam-search decoding and adjusts token selection.
\end{itemize}
Lastly, the performance of $\method$ improves in correlation with the advancement of the control guidance extractor used. 
Consequently, to demonstrate the potential upper bound of $\method$'s performance, we consider a version utilizing a ground-truth oracle extractor, which we denote as $\method$-Truth.
Further details on model architectures, datasets and evaluation metrics are deferred to Appendix~\ref{app:exp_detail}.

\paragraph{Hyperparameter setting.}
The hyperparameters for our method are fixed across tasks, with key settings including a guidance strength of 0.7, score threshold for DETR at 0.95, a detection threshold for RAM++ of 0.68, and a greedy sampling approach with a random seed of 242.

\begin{table*}[!ht]
    \centering
    \caption{Evaluation with CHAIR score across multiple LVLM architectures comparing our method with several baselines. We report CHAIR$_S$, CHAIR$_I$ and the recall score. The \textbf{bold} numbers indicate the best results among the methods evaluated and the \underline{underscored} numbers represent the second-best results. We show $\method$-Truth as a reference performance of $\method$.}
    \resizebox{\textwidth}{!}{%
\begin{tblr}{colspec = {lcccccccccccccccccc},
row{1-2} = {bg=gray!25},
row{4,6,8, 10} = {bg=gray!10},
row{8} = {bg=LightCyan},
}
    \toprule
       Method & \SetCell[c=3]{c}{LLaVA} & & & \SetCell[c=3]{c}{LLaVA-v1.5} & & & \SetCell[c=3]{c}{MiniGPTv2} & & &  \SetCell[c=3]{c}{mPLUG-Owl2} & & &  \SetCell[c=3]{c}{InstructBLIP} & & & \SetCell[c=3]{c}{\textbf{\textcolor{maroon}{Average}}} & & \\
        \cmidrule[lr]{2-4} \cmidrule[lr]{5-7} \cmidrule[lr]{8-10} \cmidrule[lr]{11-13} \cmidrule[lr]{14-16} \cmidrule[lr]{17-19} 
        \textbf{CHAIR} & $C_S \downarrow$  & $C_I \downarrow$ & $R\uparrow$ & $C_S \downarrow$  & $C_I \downarrow$ & $R\uparrow$ & $C_S \downarrow$  & $C_I \downarrow$ & $R\uparrow$ & $C_S \downarrow$  & $C_I \downarrow$ & $R\uparrow$ & $C_S \downarrow$  & $C_I \downarrow$ & $R\uparrow$ & $C_S \downarrow$  & $C_I \downarrow$ & $R\uparrow$\\
        \midrule
        Greedy & 26.6 & 10.5 & 47.4 & 8.8 & 4.6 & 41.1 & 8.2 & \underline{4.2} & 41.1 & 6.2 & {3.4} & 38.8 & {5.0} & {3.2} & 33.2 & 11.0 & 5.2 & 40.3 \\
        LURE       & 33.8         & 11.6        & \textbf{54.8} & 38.9          & 11.2           & \textbf{56.3} & 36.2          & 11.4          & \textbf{54.6} & 33.9          & 10.8          & \textbf{55.9} & 38.1          & 12.1          & \textbf{54.5}  & 36.2 & 11.4 & \textbf{55.2} \\
        Woodpecker    & {19.5} & \underline{8.9} & 44.3 & 8.5 & 4.5 & 38.4 & \underline{7.5} & 4.5 & 37.0 & 8.0 & 4.3 & 37.5 & 8.0 & 6.2 & 32.6 & 10.3 & 5.7 & 38.0 \\
        VCD & 28.1 & 11.0 & 46.6 & \underline{7.3} & \underline{4.1} & 40.8 & \textbf{6.8} & \textbf{3.9} & 38.2 & \underline{5.9} & 3.4 & 37.7 & \underline{2.4} &\underline{1.5} & 33.7 & \underline{10.1} & \underline{4.8} & 39.4 \\
        OPERA & \underline{22.4} & 9.9 & 43.6 & 11.0 & 6.7 & 40.2 & 9.2 & 5.0 & 41.3 & 5.8 & \underline{3.2} & 38.4 & 4.6 & 2.7 & \underline{38.0} & 10.6 & 5.5 & 40.3 \\
        \midrule
        \textbf{$\method$} &\textbf{17.8} & \textbf{7.2} & \underline{50.8} & \textbf{6.2} & \textbf{3.0} & 44.3 & 11.8 & 4.9 & \underline{49.7} & \textbf{4.2} & \textbf{2.3} & 41.4 & \textbf{2.2} & \textbf{1.3} & 36.3 & \textbf{8.4} & \textbf{3.7} & \underline{44.5} \\
        $\method$-Truth & 19.6 & 5.1 & 79.0 & 6.0 & 2.5 & 55.3 & 12.6 & 3.8 & 70.5 & 3.8 & 1.7 & 48.0 & 3.0 & 1.8 & 35.9 & 8.9 & 2.9 & 57.5 \\
    \bottomrule
    \end{tblr}%
    }
    \label{tab:main_chair}
\end{table*}

\begin{table*}[!ht]
    \centering
    \caption{Evaluation with POPE score in adversarial setting across multiple LVLM architectures comparing our method with several baselines. We report the POPE accuracy (\%), F1 score (\%) and the yes ratio (\%). The ideal yes ratio for a non-biased LVLM is $50\%$. The \textbf{bold} numbers indicate the best results among the methods evaluated and the \underline{underscored} numbers represent the second-best results. We show $\method$-Truth as a reference performance of $\method$.}
    \resizebox{\textwidth}{!}{%
\begin{tblr}{colspec = {lcccccccccccccccccc},
row{1-2} = {bg=gray!25},
row{4,6} = {bg=gray!10},
row{8} = {bg=LightCyan},
}
    \toprule
       Method & \SetCell[c=3]{c}{LLaVA} & & & \SetCell[c=3]{c}{LLaVA-v1.5} & & & \SetCell[c=3]{c}{MiniGPTv2} & & &  \SetCell[c=3]{c}{mPLUG-Owl2} & & &  \SetCell[c=3]{c}{InstructBLIP}  & & & \SetCell[c=3]{c}{\textbf{\textcolor{maroon}{Average}}} & & \\
        \cmidrule[lr]{2-4} \cmidrule[lr]{5-7} \cmidrule[lr]{8-10} \cmidrule[lr]{11-13} \cmidrule[lr]{14-16} \cmidrule[lr]{17-19} 
        \textbf{POPE} & Acc $\uparrow$  & F1 $\uparrow$ & Yes & Acc $\uparrow$ & F1 $\uparrow$ & Yes & Acc $\uparrow$ & F1 $\uparrow$ & Yes & Acc $\uparrow$ & F1 $\uparrow$ & Yes & Acc $\uparrow$ & F1 $\uparrow$ & Yes & Acc $\uparrow$ & F1 $\uparrow$ & Yes \\
        \midrule
        Greedy       & 51.8 & 67.4 & 97.7 & 79.4 & \underline{81.6} & 61.6 & 82.7 & 81.7 & \underline{44.5} & 72.5 & 77.5 & 72.4 & \underline{79.8} & \textbf{81.4} & 58.6 & 73.2 & 77.9 & 67.0\\
        LURE & - & - & - & - & - & - & - & - & - & - & - & - & - & - & - & - & - & - \\
        Woodpecker  & \textbf{77.5 }& \textbf{77.6} & \textbf{50.5} & \underline{80.5} & 80.6 & \textbf{50.5} & 79.5 & 77.8 & 42.5 & 77.5 & 76.9 & \underline{47.5} & 79.0 & 78.6 & \textbf{48.0} & \underline{78.8} & \underline{78.3} & \underline{47.8}\\
        VCD & 54.6 & 68.5 & 94.0 & 78.2 & 80.7 & 62.8 & 81.4 & 80.2 & 44.1 & 72.3 & 77.0 & 70.5 & 79.7 & \underline{80.9} & \underline{56.7} & 73.2 & 77.5 & 65.6 \\
        OPERA & 51.7 & 67.4 & 98.0 & 77.5 & 80.1 & 63.2 & \underline{82.9} & \underline{81.9} & 44.3 & 70.3 & \underline{79.1} & 84.6 & \underline{79.8} & \textbf{81.4} & 58.6 & 72.4 & 78.0 & 69.7\\
        \midrule
        \textbf{$\method$} & \underline{66.9} & \underline{72.9} & \underline{72.3} & \textbf{85.0} & \textbf{84.3} & \underline{45.7} & \textbf{83.0} & \textbf{82.9} & \textbf{49.4} & \textbf{82.8} & \textbf{82.7} & \textbf{49.2} & \textbf{81.7} & 79.4 & 38.8 & \textbf{79.9} & \textbf{80.4} & \textbf{51.1} \\
        $\method$-Truth & 75.6 & 80.1 & 72.3 & 92.0 & 92.5 & 57.0 & 86.9 & 88.3 & 62.5 & 93.4 & 93.8 & 56.2 & 93.8 & 93.8 & 51.0 & 88.3 &	89.7 & 59.8\\
    \bottomrule
    \end{tblr}%
    }
    \label{tab:main_pope}
\end{table*}

\begin{table}[!t]
    \centering
    \caption{Results of GPT-4V-aided evaluation. The accuracy and detailedness metrics are on a scale of 10, and a higher score indicates better performance. The symbols \(\times\) and \(\checkmark\) indicate performance metrics without and with our method, respectively.}
    \resizebox{\linewidth}{!}{%
\begin{tblr}{colspec = {llcccc},
row{1-2} = {bg=gray!25},
row{5-6} = {bg=gray!10}
}
    \toprule
    \SetCell[r=2]{l}{Task} & \SetCell[r=2]{l}{Metrics} & \SetCell[c=2]{c}{LLaVA} & & \SetCell[c=2]{c}{mPLUG-Owl2} & \\
    \cmidrule[lr]{3-4} \cmidrule[lr]{5-6}
    & & \XSolidBrush & \checkmark & \XSolidBrush & \checkmark \\
    \midrule
    \SetCell[r=2]{l}{LLaVA-QA90} & Acc $\uparrow$ & $5.82_{\pm 0.10}$ & \textbf{5.94$_{\pm 0.05}$} & $6.03_{\pm 0.13}$ & \textbf{6.35$_{\pm 0.21}$} \\
    & Detail $\uparrow$ & $4.59_{\pm 0.08}$ & {4.59$_{\pm 0.08}$} & $5.06_{\pm 0.05}$ & \textbf{5.16$_{\pm 0.10}$} \\
    \SetCell[r=2]{l}{Image Captioning} & Acc $\uparrow$ & $5.27_{\pm 0.20}$ & \textbf{6.11$_{\pm 0.23}$} & $7.97_{\pm 0.25}$ & \textbf{8.63$_{\pm 0.20}$} \\
    & Detail $\uparrow$ & \textbf{4.39$_{\pm 0.29}$} & {4.36$_{\pm 0.17}$} & $5.74_{\pm 0.24}$ & \textbf{6.19$_{\pm 0.23}$} \\
    \bottomrule
    \end{tblr}%
    }
    \label{tab:gpt4v-aided}
\end{table}

\subsection{Results}
Experimental results on object hallucination metrics (CHAIR and POPE) are presented in Table~\ref{tab:main_chair} and ~\ref{tab:main_pope}. 
Overall, $\method$ achieves superior performances across different LVLM architectures and evaluation metrics.

\paragraph{Results on CHAIR.} 
CHAIR is a widely adopted benchmark for evaluating caption hallucination in LVLMs, comparing generated descriptions with ground-truth object annotations. 
It captures object-level precision through CHAIR$_I$ (instance-level) and CHAIR$_S$ (sentence-level), and we further report Recall to assess content coverage.

Table~\ref{tab:main_chair} shows that $\method$ consistently outperforms existing approaches on all major metrics. 
It achieves the lowest average CHAIR$_I$ and CHAIR$_S$ scores and ranks second in Recall, reducing hallucination without sacrificing coverage.
Compared to the second-best method, $\method$ improves CHAIR$_S$ by $1.7$ points and CHAIR$_I$ by $1.1$ on average.
The gains are particularly strong on LLaVA models, where hallucination drops by up to $8.8$ points. 
In contrast, methods such as LURE and Woodpecker are less effective across model variants.

Importantly, $\method$ achieves performance comparable to $\method$ -Truth, a variant that uses ground-truth object labels as guidance. 
This finding suggests that aggregating signals from multiple visual models offers a compelling alternative to manual supervision to reduce hallucination.

\paragraph{Results on POPE.} 
POPE is designed to assess object-level grounding in LVLMs by testing their ability to answer yes/no questions about visual content. 
We focus on the adversarial setting, which presents challenging negatives and helps expose hallucination and biased answering tendencies.

In Table~\ref{tab:main_pope}, $\method$ consistently outperforms all baselines, with average improvements of $6.7\%$ in accuracy and $3.5\%$ in F1 score over the original model outputs. 
Compared to the second-best method, Woodpecker, $\method$ still maintains a $1.1\%$ gain in accuracy and a $2.1\%$ gain in F1.

Beyond accuracy, $\method$ also reduces the overconfident bias often seen in LVLMs’ outputs. 
This is reflected in a more balanced ``yes'' ratio (closer to $50\%$, reflecting a $15.9\%$ shift towards unbiased answers). This shift suggests that $\method$ produces more trustworthy predictions by reducing the tendency toward overconfident affirmative responses.

\paragraph{Results on GPT-4V-aided evaluation.}
Following~\citet{yin2023woodpecker}, this GPT-4V-assisted evaluation provides a qualitative perspective that complements the numerical metrics of CHAIR and POPE, offering a more comprehensive assessment of model performance. 
As shown in Table~\ref{tab:gpt4v-aided}, GPT-4V consistently assigns higher accuracy with equal detailedness scores to models enhanced by $\method$, highlighting its ability to produce more precise and detailed descriptions, which demonstrates the robustness of our method in real-world visual tasks. 
The evaluation prompt is detailed in Appendix~\ref{app:setting_hallu}.

\begin{figure}[!ht]
    \centering
    \includegraphics[width=0.9\linewidth]{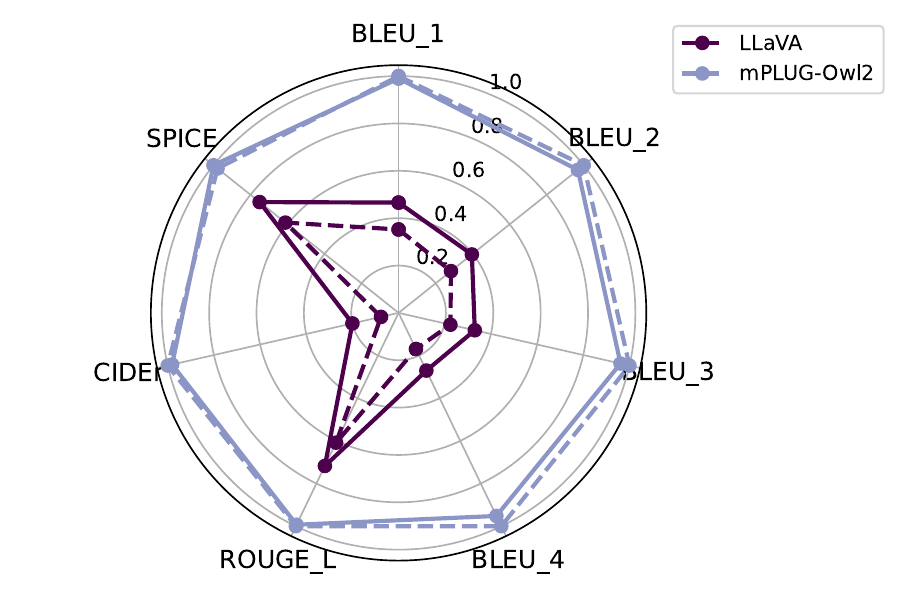}
    \caption{
    $\method$ maintains or improves overall text quality on general metrics. Solid lines indicate models with $\method$, while dashed lines indicate the original models. Higher scores indicate better textual similarity to the reference outputs.}
    \label{fig:radar_general}
\end{figure}

\paragraph{Additional results on other vision-language tasks.}
To further evaluate the generalizability of our approach beyond object hallucination and the MSCOCO dataset, we extended our evaluations to additional datasets including A-OKVQA and GQA and included more general caption quality metrics. 
As shown in Table~\ref{tab:additional_pope_avg}, the POPE results demonstrate that our method consistently mitigates hallucinations across various datasets with different image distributions. 
Figure~\ref{fig:radar_general} presents a comprehensive evaluation of the image captioning task on MSCOCO and LLaVA-QA90, a comprehensive VQA dataset, using metrics including BLEU~\citep{papineni2002bleu}, ROUGE~\citep{lin-2004-rouge}, CIDEr~\citep{vedantam2015cider} and SPICE~\citep{anderson2016spice}. 
These results demonstrate that, although our method primarily targets hallucination mitigation, it maintains the overall performance of LVLMs on broader tasks, with no significant trade-offs in caption or VQA quality.

\begin{table}[!t]
    \centering
    \caption{POPE results across three datasets. We report the average score under random, popular, adversarial settings. The detailed POPE results can be found in the appendix~\ref{app:exp_results}. The \textbf{bold} numbers indicate the best results. The ideal yes ratio for a non-biased LVLM is $50\%$.}
    \resizebox{\linewidth}{!}{%
\begin{tblr}{colspec = {lccccccc},
row{1-2} = {bg=gray!25},
row{5-6} = {bg=gray!10}
}
    \toprule
    \SetCell[r=2]{l}{Dataset} & \SetCell[r=2]{c}{w/$\method$} & \SetCell[c=3]{c}{LLaVA} & & & \SetCell[c=3]{c}{mPLUG-Owl2} & \\
    \cmidrule[lr]{3-5} \cmidrule[lr]{6-8}
    &  & Accuracy $\uparrow$ & F1 $\uparrow$ & Yes(\%) & Accuracy $\uparrow$ & F1 $\uparrow$ & Yes(\%) \\
    \midrule
    \SetCell[r=2]{l}{MSCOCO} & \XSolidBrush  & 54.2 & 68.5 & 95.5 & 76.7 & 80.4 & 68.2 \\
    & \checkmark & \textbf{72.2} & \textbf{76.4} & \textbf{66.9} & \textbf{85.5} & \textbf{85.0} & \textbf{46.5} \\
    \SetCell[r=2]{l}{A-OKVQA} & \XSolidBrush  & 51.8 & 67.5 & 97.9 & 69.6 & 76.5 & 78.5 \\
    & \checkmark & \textbf{64.3} & \textbf{72.8} & \textbf{80.2} & \textbf{82.0} & \textbf{83.5} & \textbf{57.2} \\
    \SetCell[r=2]{l}{GQA} & \XSolidBrush  & 52.0 & 67.6 & 97.8 & 73.7 & 78.7 & 72.6 \\
    & \checkmark & \textbf{62.5} & \textbf{71.8} & \textbf{81.8} & \textbf{80.1} & \textbf{80.6} & \textbf{51.1} \\
    \bottomrule
    \end{tblr}%
    }
    \label{tab:additional_pope_avg}
\end{table}

\textbf{Latency analysis}
Many existing approaches to mitigating object hallucination rely on post-generation correction models~\citep{zhou2023analyzing, zhai2023halle, yin2023woodpecker}, external object detectors~\citep{yin2023woodpecker}, or complex decoding strategies~\citep{huang2023opera, leng2023mitigating}, all of which introduce substantial computational overhead. 
To assess the practical efficiency of $\method$, we evaluate its latency compared to existing baselines on LLaVA-7B, as shown in Table~\ref{tab:latency}.

Our measurements include the time required for additional forward passes through external vision models. These models contribute only marginal latency relative to the cost of autoregressive decoding in LVLMs. 
In general, $\method$ increases decoding time by just 1.98$\times$, the lowest among all baselines. 
This demonstrates that $\method$ achieves the most favorable trade-off between latency and accuracy, which makes it suitable for real-world use. Detailed settings are provided in Appendix~\ref{app:setting_other}.

\begin{table*}[!t]
    \centering
    \caption{Inference latency comparison. We report both the latency and the ratio to the latency of greedy decoding of the original LVLM model.}
    \resizebox{0.9\textwidth}{!}{%
\begin{tblr}{colspec = {lcccccc},
row{1} = {bg=gray!25},
row{3} = {bg=gray!10}
}
    \toprule 
    & Greedy & LURE & Woodpecker$^*$ & VCD & OPERA & \textbf{$\method$} (ours) \\
    \midrule
    Training Cost & 0 & 10min on A100 80G & 0 & 0 & 0 & 0 \\
    Inference Latency$^{\text{(ms/token)}}$ & 26.3 (\small{×1.0}) & 179.9 (\small{×6.84}) & 94.5 (\small{×3.59})* & 53.4 (\small{×2.03}) & 185.1 (\small{×7.0}) & \textbf{52.2} (\small{×\textbf{1.98}}) \\
    \bottomrule
    \end{tblr}%
    }
    \scriptsize{\\$^*$Woodpecker requires GPT API key access and the latency may depend on OPENAI API. \hfill}
    \label{tab:latency}
\end{table*}

\subsection{Ablation Study}\label{sec:ablation}
\paragraph{Why incorporate multiple image-grounded models?}
Different image-grounded models excel at capturing different aspects of visual information—some detect objects precisely, while others offer broader, fine-grained context. 
To understand whether combining these complementary signals leads to better guidance, we conduct an ablation comparing DETR and RAM++ individually versus in combination (Table~\ref{tab:ablation_ensembling}). 
All variants are evaluated under the same decoding setup to ensure a fair comparison.

DETR allows for highly accurate object detection, while RAM++ excels in extensive recognition tasks, contributing fine-grained visual concepts.
Their combination yields consistent improvements on CHAIR metrics, suggesting that aggregating multiple visual perspectives is important for effective hallucination mitigation.

\paragraph{What is the best way to integrate guidance from multiple models?}
When aggregating the outputs from multiple image-grounding models, the combination method can significantly affect guidance quality. 
We compare two strategies: taking the intersection or the union of detected objects.

As shown in Table~\ref{tab:ablation_methods_ensemble}, the intersection-based approach consistently outperforms the union, significantly reducing hallucination. 
This suggests that enforcing agreement across models leads to more precise and trustworthy guidance, while union-based aggregation may introduce noisy or spurious information. 
The detailed experimental setup and prompt templates are provided in Appendix~\ref{app:exp_detail}.

\begin{table}[!t]
    \centering
    \caption{Ablation study comparing the performance of combining DETR and RAM++ models versus using individual vision models. This approach leverages multiple object detectors to provide more reliable and robust object-level guidance, resulting in superior performance on CHAIR metrics.}
    \resizebox{\linewidth}{!}{%
    \begin{tblr}{colspec = {lcccccc},
    row{1-2} = {bg=gray!25},
    row{4-5} = {bg=gray!10}
    }
        \toprule 
        \textbf{Model} & \SetCell[c=2]{c}{\textbf{LLaVA}} &  & \SetCell[c=2]{c}{\textbf{LLaVA-v1.5}} & & \SetCell[c=2]{c}{\textbf{mPLUG-Owl2}} \\
        \cmidrule[lr]{2-3} \cmidrule[lr]{4-5} \cmidrule[lr]{6-7}
        \textbf{CHAIR} & $C_S \downarrow$ & $C_I  \downarrow$ & $C_S \downarrow$ & $C_I  \downarrow$ & $C_S \downarrow$ & $C_I  \downarrow$ \\
        \midrule
        Greedy & 26.6 & 10.5 & 8.8 & 4.6 & 6.2 & 3.4\\

        \textit{Ensembling Models} \\
        \texttt{$\method$} & \textbf{17.8} & \textbf{7.2}  & \textbf{6.2}  & \textbf{3.0}  & \textbf{4.2}  & \textbf{2.3}   \\
        \textit{Single Models} \\
        \texttt{$\method$-DETR only}           & 27.6 & 8.4 & 10.5 & 4.3 & 5.3 & 2.7 \\
        \texttt{$\method$-RAM only}            & 29.0 & 9.1  & 6.6   & 3.7  & 5.2  & 2.8   \\
        \bottomrule
    \end{tblr}%
    }
    \label{tab:ablation_ensembling}
\end{table}

\begin{table}[!t]
    \centering
    \caption{Effect of Integration Methods for Image-Grounding Models.}
    \resizebox{\linewidth}{!}{%
    \begin{tblr}{colspec = {lcccccc},
    row{1-2} = {bg=gray!25},
    row{4} = {bg=gray!10}
    }
        \toprule 
        \textbf{Model} & \SetCell[c=2]{c}{\textbf{LLaVA}} &  & \SetCell[c=2]{c}{\textbf{LLaVA-v1.5}} & & \SetCell[c=2]{c}{\textbf{mPLUG-Owl2}} \\
        \cmidrule[lr]{2-3} \cmidrule[lr]{4-5} \cmidrule[lr]{6-7}
        \textbf{CHAIR} & $C_S \downarrow$ & $C_I  \downarrow$ & $C_S \downarrow$ & $C_I  \downarrow$ & $C_S \downarrow$ & $C_I  \downarrow$ \\
        \midrule
        Greedy & 26.6 & 10.5 & 8.8 & 4.6 & 6.2 & 3.4\\
        \texttt{$\method$-intersection} (ours) & \textbf{17.8} & \textbf{7.2}  & 6.2  & 3.0 & \textbf{4.2}  & \textbf{2.3}  \\
        \texttt{$\method$-union} & 30.4 & 9.7  & \textbf{5.4}  & \textbf{2.7}  & 4.8  & 2.7 \\
        \bottomrule
    \end{tblr}%
    }
    \label{tab:ablation_methods_ensemble}
\end{table}

\paragraph{How does control strength affect generation?}
To understand the impact of guidance strength in our decoding setup, we vary the control weight $\gamma$, which balances the influence between the original LVLM generation and the generation conditioned on external image-grounded guidance. 

\begin{figure}[t!]
    \centering
    \includegraphics[width=\linewidth]{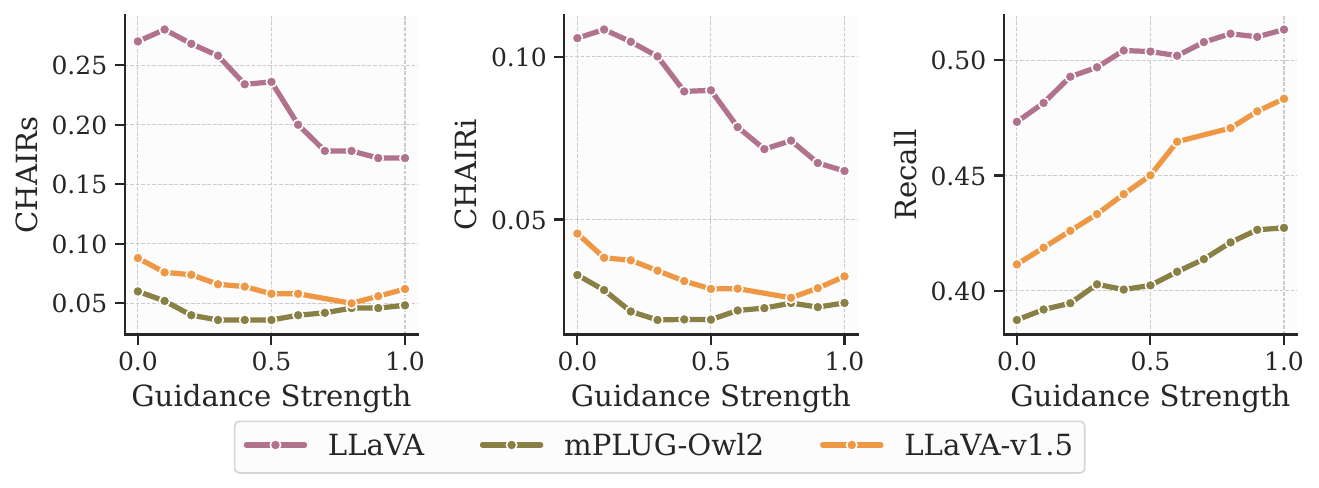}
    \vspace{-1em}
    \caption{Ablation study on the effect of guidance strength ($\gamma$) on the performance of LLaVA, LLaVA-v1.5 and mPLUG-Owl2 using CHAIR metrics, with $\gamma$ ranging from 0 to 1.}
\label{fig:ablation_llava}
\end{figure}

Figure~\ref{fig:ablation_llava} shows that increasing guidance strength from $0$ to $1$ leads to a notable decrease in CHAIR scores. 
This trend suggests that higher guidance strength makes LVLMs rely more on image-grounded features, thereby enhancing their ability to produce accurate descriptions. It's crucial to note that, although some models exhibit optimal performance at a guidance strength of $\gamma=1$, excessively strong guidance can adversely affect the models' ability to adhere to provided instructions. Experimental evidence is detailed in Appendix~\ref{app:ab_guidance_strength}.
This observation highlights the necessity of having a balanced guidance strength that ensures high-quality, accurate outputs while adhering closely to the given instructions. 
Based on our findings, we recommend a guidance strength within the range of $\gamma \in (0.3,0.7)$ as the most effective for maintaining this balance.

\begin{figure}[t!]
    \centering
    \includegraphics[width=\linewidth]{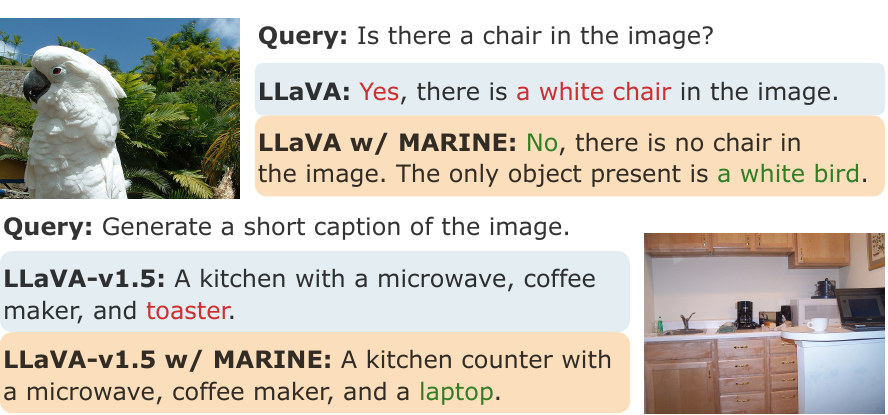}
    \caption{Hallucination mitigation examples by our proposed $\method$ across multiple tasks. Hallucinated objects generated by the LVLM are highlighted in red.}
    \label{fig:examples}
\end{figure}

\section{Conclusions, Limitations and Future Work}\label{sec:conc}
In this paper, we introduced a training-free and API-free framework $\method$ to mitigate object hallucination in LVLMs during its text generation process. 
Leveraging a pre-trained object grounding vision encoder for a novel guidance framework in the multi-modal setting, $\method$ effectively and cost-efficiently reduces the hallucinations of five widely-used LVLMs, as assessed by various metrics across different tasks. 
The inherent compatibility of the $\method$ with various vision models and projection functions further underscores its flexibility.  
In contrast to post-generation correction methods, $\method$ strikes a balance between efficiency, instruction-following ability and effectiveness in reducing object hallucinations. 

\noindent\textbf{Limitations and future work.} While $\method$ has demonstrated impressive performance by utilizing guidance from image-grounded models, there remains potential for further improvement through the integration of advanced aggregation methods, such as multi-agent debate~\citep{du2023improving}, into the~\method~framework. Additionally, although~\method~is specifically designed to mitigate object hallucination, which is the most significant issue in LVLMs, extending its application to address other types of hallucinations in both LLMs and LVLMs across a broader range of benchmarks would be highly advantageous.

\section*{Acknowledgments}
We thank anonymous reviewers for their helpful comments. Part of this work was done while WZ was a PhD student at UCLA. WZ and QG are supported in part by NSF grants DMS-2323113, CPS-2312094, IIS-2403400, and the research fund from the UCLA-Amazon Science Hub. WZ was also supported by the UCLA dissertation year fellowship. The views and conclusions contained in this paper are those of the authors and should not be interpreted as representing any funding agencies.

\section*{Impact Statement}
This paper introduces research aimed at advancing the field of Large Language Models. We are confident that our work will contribute to significant social benefits, particularly by enhancing the accountability of LLMs through the reduction of hallucinatory outputs. Our proposed method, \method, holds the potential to improve the fairness of LLM interactions by effectively reducing biased hallucinations. By mitigating hallucinations, \method~has the potential to offer a positive social impact by ensuring that LVLMs generate more accountable responses. Despite this merit, \method~cannot address prejudicial biases inherent in LLM prior knowledge, which could be a focus of future work. To the best of our knowledge, we have not identified any negative effects associated with our research that merit highlighting in this discussion.

\bibliography{main}
\bibliographystyle{icml2025}
\clearpage
\appendix
\onecolumn

\section{Experiment Setup}\label{app:exp_detail}
We conduct all of the experiments using 8 A6000 GPU with 48GB GPU memory. Each single experiment can be run on a single A6000 GPU. 

\subsection{Model Architectures}\label{app:model_arch}
In Table~\ref{tab:app_arch}, we provide detailed descriptions of the LVLM architectures used in our experiments. These LVLMs respectively leverage the pre-trained vision encoder of the models we listed, which are all based on the Vision Transformer (ViT)~\citep{dosovitskiy2020image} architecture. 
\begin{table}[ht!]
    \centering
    \caption{Details of the LVLM architectures that we used in our paper.}
    \resizebox{0.99\textwidth}{!}{%
    \begin{tabular}{c|c c}
        \toprule
         Model & Vision encoder & LLM \\
         \midrule
         LLaVA~\citep{liu2023llava} & CLIP-L~\citep{radford2021learning} & LLaMA-2-7B-Chat~\citep{touvron2023llama2} \\
         LLaVA-v1.5~\citep{liu2023improved} & CLIP-L-336px~\citep{radford2021learning} & Vicuna-v1.5-7B~\citep{vicuna2023} \\
         MiniGPT-v2~\citep{chen2023minigpt} & EVA-G~\citep{fang2023eva}  & LLaMA-2-7B-Chat~\citep{touvron2023llama2}\\
         mPLUG-OWL2~\citep{ye2023mplug} &  CLIP-L~\citep{radford2021learning} & LLaMA-2-7B~\citep{touvron2023llama2} \\
         InstructBLIP~\citep{dai2023instructblip} & BLIP-2~\citep{li2023blip} & Vicuna-v1.1-7B~\citep{vicuna2023} \\
         \bottomrule
    \end{tabular}%
    }
    \label{tab:app_arch}
\end{table}

\subsection{Descriptions about Additional Metrics}\label{app:exp_metrics}
In Figure~\ref{fig:radar_general}, we evaluate the text quality of the outputs generated with $\method$ using general metrics as follows:
\begin{itemize}[leftmargin=*,nosep]
    \item \emph{BLEU}~\citep{papineni2002bleu} measures how well the generated translation matches the reference translations in
    terms of n-gram overlap.
    \item \emph{ROUGE-L}~\citep{lin-2004-rouge} measures the quality of a machine-generated summary by comparing it to one or more reference summaries.
    \item \emph{CIDEr}~\citep{vedantam2015cider} assesses the quality of image captioning models. It focuses on evaluating how well the
    generated captions align with human consensus.
    \item \emph{SPICE}~\citep{anderson2016spice}  focuses on assessing the semantic similarity between the generated captions and reference captions.
\end{itemize}

\subsection{Prompt Templates}
For each query, we randomly select a prompt template from the available template list, as shown in Table~\ref{tab:prompt_template}.
\begin{table}[ht!]
    \centering
    \caption{Details of the LVLM architectures that we used in our paper.}
    \resizebox{0.9\textwidth}{!}{%
    \begin{tabular}{p{3cm}|p{12cm}}
    \hline
    \textbf{Template Type} & \textbf{Prompt Template} \\
    \hline
    {$\method$-intersec} & This image contains \texttt{<OBJECT\_GROUNDING>}. Based on this, \texttt{<QUERY>} \\
    
    & The image contains the following objects: \texttt{<OBJECT\_GROUNDING>}. Given these detected objects, \texttt{<QUERY>} \\
    
    & This image shows the following objects: \texttt{<OBJECT\_GROUNDING>}. Using this information, \texttt{<QUERY>} \\
    
    & The objects found in this image are: \texttt{<OBJECT\_GROUNDING>}. Considering this list of objects, \texttt{<QUERY>} \\
    \hline

    {POPE task} & This image contains only the following objects: \texttt{<OBJECT\_GROUNDING>}. Do not assume anything beyond these objects. Based solely on this list, \texttt{<QUERY>} \\

    & The detected objects in the image are: \texttt{<OBJECT\_GROUNDING>}. Answer based only on these objects. \texttt{<QUERY>} \\

    & This image shows the following objects: \texttt{<OBJECT\_GROUNDING>}. You must answer using only the objects in this list. Given these detected objects, \texttt{<QUERY>} \\

    & The objects found in this image are limited to: \texttt{<OBJECT\_GROUNDING>}. You should rely strictly on this list of objects and make no other guesses. Based on this, \texttt{<QUERY>} \\
    \hline
    {$\method$-union} & List of detected objects in the image: \\
    & \texttt{<OBJECT\_GROUNDING\_A>} \\
    & \texttt{<OBJECT\_GROUNDING\_B>} \\
    & Based on the detected objects above, \texttt{<QUERY>} \\

    & The most prominent objects detected are: \\
    & \texttt{<OBJECT\_GROUNDING\_A>} \\
    & \texttt{<OBJECT\_GROUNDING\_B>} \\
    & Given these findings, \texttt{<QUERY>} \\

    & The following objects were detected in the image: \\
    & \texttt{<OBJECT\_GROUNDING\_A>} \\
    & \texttt{<OBJECT\_GROUNDING\_B>} \\
    & With this information, \texttt{<QUERY>} \\

    & Here is a list of all objects detected in the image: \\
    & \texttt{<OBJECT\_GROUNDING\_A>} \\
    & \texttt{<OBJECT\_GROUNDING\_B>} \\
    & Do not infer or hallucinate any additional objects. Using only the detected objects, \texttt{<QUERY>} \\
    \hline
    \end{tabular}%
    }
    \label{tab:prompt_template}
\end{table}

\subsection{Details of Baselines}\label{app:exp_baselines}
Specifically, the hyperparameters for LURE~\citep{zhou2023analyzing}, VCD~\citep{leng2023mitigating}, OPERA~\citep{huang2023opera} are reported in Table~\ref{tab:lure_hyperparameters}, ~\ref{tab:vcd_hyperparameters} and ~\ref{tab:opera_hyperparameters} respectively. We strictly followed the original implementations and default hyperparameters described in their papers to reproduce the results for each baseline.

\begin{table}[ht!]
\centering
\caption{LURE~\citep{zhou2023analyzing} Hyperparameter Settings}
\label{tab:lure_hyperparameters}
\begin{tabular}{l|c}
\hline
Parameters & Value \\
\hline
Uncertainty Threshold $\gamma$ & 0.9 \\
Position Threshold $\iota$ & 0.8 \\
\hline
\end{tabular}
\end{table}

\begin{table}[ht!]
\centering
\caption{VCD~\citep{leng2023mitigating} Hyperparameter Settings}
\label{tab:vcd_hyperparameters}
\begin{tabular}{l|c}
\hline
Parameters & Value \\
\hline
Amplification Factor $\alpha$ & 1 \\
Adaptive Plausibility Threshold & 0.1 \\
Diffusion Noise Step & 500 \\
\hline
\end{tabular}
\end{table}

\begin{table}[ht!]
\centering
\caption{OPERA~\citep{huang2023opera} Hyperparameter Settings}
\label{tab:opera_hyperparameters}
\begin{tabular}{lc}
\hline
Parameters & Value \\
\hline
Self-attention Weights Scale Factor $\theta$ & 50 \\
Attending Retrospection Threshold & 25 \\
Beam Size & 5 \\
Attention Candidates & 1 \\
Penalty Weights & 1\\
\hline
\end{tabular}
\end{table}

\begin{table}[ht!]
\centering
\caption{$\method$ Hyperparameter Settings. The settings are fixed depending on the question-answering tasks.}
\label{tab:MARINE_settings}
\begin{tabular}{l c}
\hline
\textbf{Parameters} & \textbf{Value} \\
\hline
\multicolumn{2}{l}{\textit{Guidance}} \\
Guidance Strength & 0.7 \\
score threshold for DETR & 0.95 \\
Detect Threshold for RAM++ & 0.68 \\
\hline
\multicolumn{2}{l}{\textit{Generation}} \\
Max Token Length & 64 \\
Sampling & Greedy \\
Random Seed & 242 \\
\hline
\end{tabular}
\end{table}

\begin{table}[ht!]
\caption{Batch size for LVLM generation is fixed across all experiments unless otherwise noted. To expedite the evaluation process, we employed the batched generation. We avoid the negative impact of batched generation by adopting left padding if the LVLM does not explicitly assign the padding strategy for inference.}
\label{tab:inference_setting}
\begin{center}
\resizebox{0.65\textwidth}{!}{%
\begin{tabular}{lccccc}
\toprule
 Model & LLaVA & LLaVA-v1.5 & MiniGPTv & mPLUG-Owl2 & InstructBLIP \\
\midrule
Batch Size & 16 & 4 & 32 & 16 & 16 \\
\bottomrule
\end{tabular}%
}
\end{center}
\end{table}

\subsection{Experiment Setting for Hallucination Evaluations}\label{app:setting_hallu}
Key factors that potentially affect the hallucination evaluation outcomes, including the evaluation dataset and prompt template, LVLM's sampling strategy and batched generation techniques, and guidance strength, are detailed in this section. The hyper-parameters setting for $\method$ and overall experiment settings are shown in Table~\ref{tab:MARINE_settings} and ~\ref{tab:inference_setting}.

\textbf{Experiment setting for CHAIR evaluation.} 
We adopt the same prompt ``Generate a short caption of the image.'' as utilized by ~\citet{li2023evaluating}. 
The hyperparameters are fixed, including a guidance strength of 0.7, score threshold for DETR at 0.95, a detection threshold for RAM++ of 0.68, a maximum token length of 64, and a greedy sampling approach with a random seed of 242.\\
For the calculation of CHAIR metrics, we referenced the 80 object categories annotated in the MSCOCO dataset, following ~\citet{rohrbach2018object}. Besides, we employed the synonym list from ~\citet{lu2018} to align synonymous words in the generated text with MSCOCO object categories. Additionally, due to the cost considerations associated with the GPT-3.5 API, we limited our analysis to 200 samples for Woodpecker correction for each model and reported the result in Table~\ref{tab:main_chair}.

\textbf{Experiment setting for POPE evaluation.} 
POPE is a flexible approach to evaluating hallucinations in LVLMs, which formulates a binary classification task by prompting LVLMs with questions such as ``Is there a keyboard in this image?'' to answer ``yes'' or ``no''. 
Following ~\citet{li2023evaluating}, we created 3000 POPE questions across three datasets—500 images each from MSCOCO, A-OKVQA, and GQA for the POPE evaluation. 
We reported the adversarial settings in Table~\ref{tab:main_pope}, the most challenging setting, which constructs POPE questions from the top-k most frequently co-occurring but absent objects. Additionally, in Table~\ref{tab:additional_pope_avg}, we reported the average scores under random, popular,
adversarial settings across MSCOCO, A-OKVQA, and GQA datasets. The full POPE results are in Tabel~\ref{tab:additional_pope_detailed}.

Similarly, we constrained our analysis to 200 samples for Woodpecker correction for each model due to the high costs associated with the GPT API. The outcomes of this analysis are detailed in Table~\ref{tab:main_pope}.

\textbf{Experiment setting for GPT-4V-aided evaluation.} 
The GPT-4V-aided evaluation compares the outputs of two LVLM assistants using GPT-4V as a judge. We prompted GPT-4V to assess the quality of the generated outputs, scoring them out of 10 in two aspects: 
\begin{itemize}[leftmargin=*,nosep]
    \item \emph{Accuracy}: how accurately each assistant describes the image;  
    \item \emph{Detailedness}: the richness of necessary details in the response. 
\end{itemize}

As shown in Table~\ref{tab:prompt-template}, the assessment prompt template we used is slightly different from that of~\citet{yin2023woodpecker}. Specifically, we include the original question for a task-orientated evaluation and exclude prompts that describe Woodpecker-specific output formats like object bounding boxes. 

\textbf{Experiment setting for ablation study.} 
\label{app:setting_ablation}
To explore different methods of integrating image-grounding models, we investigate the intersection and union of detected objects, with integration based on synonyms using the NLTK package.\\
To quantitatively assess the influence of guidance strength, we varied it from 0 to 1, as shown in Figure~\ref{fig:ablation_logits_alternation}. These quantitative experiments were conducted using the same setting as those in CHAIR evaluation.
For qualitative analysis, we selected guidance strength from a recommended range of $\gamma \in (0.3,0.7)$.

\begin{table}[h!]
\centering
\caption{Prompt template for GPT-4V-aided evaluation. \{question\} is the original instruction; \{answer 1\} is the original response, and \{answer 2\} is the response generated by the LVLM using $\method$.}
\label{tab:prompt-template}
\begin{tabular}{p{0.8\linewidth}}
\toprule
\textbf{Prompt template for GPT-4V-aided evaluation} \\
\midrule
You are required to score the performance of two AI assistants in describing a given image. You should pay extra attention to the hallucination, which refers to the part of descriptions that are inconsistent with the image content, such as claiming the existence of something not present in the image. \\
\\
Please rate the responses of the assistants on a scale of 1 to 10, where a higher score indicates better performance, according to the following criteria: \\
1. {Accuracy}: whether the response is accurate with respect to the image content. Responses with fewer hallucinations should be given higher scores. \\
2. {Detailedness}: whether the response is rich in necessary details. Note that hallucinated descriptions should not count as necessary details. \\
\\
Please output a single line for each criterion, containing only two values indicating the scores for Assistant 1 and 2, respectively. The two scores are separated by a space. Following the scores, please provide an explanation of your evaluation, avoiding any potential bias and ensuring that the order in which the responses were presented does not affect your judgment. \\
\\
{Question:} \{\texttt{question}\} \\
{Assistant 1:} \{\texttt{answer 1}\} \\
{Assistant 2:} \{\texttt{answer 2}\} \\
\\
{Output format:} \\
Accuracy: \\
Scores of the two answers: \\
Reason: \\
Detailedness: \\
Scores of the two answers: \\
Reason: \\
\bottomrule
\end{tabular}
\end{table}

\subsection{Experiment Setting on Other Vision-Language Tasks}\label{app:setting_other}

\textbf{Experiment setting for text quality analysis.} 
For text quality analysis, we adopted 90 visual questions from the LLaVA-QA90
\footnote{\footnotesize \url{https://github.com/haotian-liu/LLaVA/blob/main/playground/data/coco2014_val_gpt4_qa_30x3.jsonl}}
task (including conversations, visual perceptions, and complex reasoning subtasks), and randomly selected 500 MSCOCO images for image captioning task. Following ~\citet{liu2023llava}, we adpoted the response generated by text-only GPT-4 (0314) with the context captions/boxes provided.
answers given by GPT-4 as references for LLaVA-QA90 task and used image captions provided in MSCOCO annotations as references for image captioning task. 

In Table~\ref{tab:ap_general_caption} and Table~\ref{tab:ap_general_qa90}, we present a detailed evaluation on the image captioning task for both MSCOCO and LLaVA-QA90 using metrics including BLEU, ROUGE, CIDEr and SPICE. The corresponding
figure result is shown in Figure~\ref{fig:radar_general}.

\textbf{Experiment setting for latency analysis.} 
We compared our method with existing baselines in terms of the trade-off between inference cost and the effectiveness of reducing object hallucinations, as shown in Table~\ref{tab:latency}.
For post-correction baselines such as Woodpecker and LURE, we first prompted LLaVA (\texttt{llava-llama-2-7b-chat-lightning-preview}) to generate captions and then measure the latency of generating the corrected outputs. The total latency for post-correction baselines includes both the generation and correction processes.
For decoding methods such as VCD, OPERA and our method, we measured the latency of LLaVA generating captions directly.

We prompted the models with ``Generate a short caption of the image.'' on 500 MSCOCO images with a batch size of 1 and a maximum token length of 64, without any stopping criteria, using a single A6000 GPU. Then latency was calculated as the ratio of the number of output tokens and encoding and generation time.

\begin{table*}[ht!]
\caption{Detailed POPE~\citep{li2023evaluating} results on three datasets (MSCOCO~\citep{lin2014microsoft}, A-OKVQA~\citep{schwenk2022okvqa}, GQA~\citep{hudson2019gqa}).}
\label{tab:additional_pope_detailed}
\centering
\resizebox{0.95\textwidth}{!}{%
\begin{tabular}{lllcccccc}
\toprule
Dataset & Type & Model & {w/$\method$} & Accuracy $\uparrow$ & Precision $\uparrow$ & Recall $\uparrow$ & F1 $\uparrow$ & Yes(\%) \\
\midrule
\multirow{12}{*}{MSCOCO} 
    & \multirow{4}{*}{Adversarial} 
        & \multirow{2}{*}{LLaVA} & \XSolidBrush & 51.8 & 50.9 & 99.5 & 67.4 & 97.7 \\
        && & \checkmark & 66.9 & 61.7 & 89.1 & 72.9 & 72.3 \\
        && \multirow{2}{*}{mPLUG-Owl2} & \XSolidBrush &
        72.5 & 65.5 & 94.9 & 77.5 & 72.4 \\
        && & \checkmark & 82.8 & 83.4 & 82.0 & 82.7 & 49.2 \\
    & \multirow{4}{*}{Popular} 
        & \multirow{2}{*}{LLaVA} & \XSolidBrush & 52.4 & 51.2 & 99.8 & 67.7 & 97.4 \\
         && & \checkmark & 71.3 & 65.8 & 88.9 & 75.6 & 67.5 \\
        && \multirow{2}{*}{mPLUG-Owl2} & \XSolidBrush & 75.8 & 68.7 & 94.9 & 79.7 & 69.0 \\
        && & \checkmark & 85.6 & 88.4 & 82.0 & 85.1 & 46.4 \\
    & \multirow{4}{*}{Random} 
        & \multirow{2}{*}{LLaVA} & \XSolidBrush & 58.3 & 54.5 & 99.7 & 70.5 & 91.4 \\
        && & \checkmark & 78.5 & 73.4 & 89.3 & 80.6 & 60.8 \\
        && \multirow{2}{*}{mPLUG-Owl2} & \XSolidBrush & 81.8 & 75.2 & 94.9 & 83.9 & 63.1 \\
        && & \checkmark & 88.1 & 93.4 & 81.9 & 87.3 & 43.9 \\
\midrule
\multirow{12}{*}{A-OKVQA} 
    & \multirow{4}{*}{Adversial} 
        & \multirow{2}{*}{LLaVA} & \XSolidBrush & 50.0 & 50.0 & 99.5 & 66.6 & 99.5 \\
        && & \checkmark & 56.3 & 53.6 & 94.3 & 68.3 & 88.1 \\
        && \multirow{2}{*}{mPLUG-Owl2} & \XSolidBrush 
        & 62.5 & 57.3 & 98.1 & 72.3 & 85.6 \\

        && & \checkmark & 74.4 & 68.8 & 89.3 & 77.7 & 64.9 \\
    & \multirow{4}{*}{Popular} 
        & \multirow{2}{*}{LLaVA} & \XSolidBrush & 50.1 & 50.1 & 99.8 & 66.7 & 99.7 \\
        && & \checkmark & 63.0 & 58.0 & 94.5 & 71.9 & 81.6 \\
        && \multirow{2}{*}{mPLUG-Owl2} & \XSolidBrush &69.1 & 62.1 & 97.9 & 76.0 & 78.9 \\

        && & \checkmark & 82.5 & 78.8 & 89.1 & 83.6 & 56.5 \\
    & \multirow{4}{*}{Random} 
        & \multirow{2}{*}{LLaVA} & \XSolidBrush & 55.4 & 52.8 & 99.8 & 69.1 & 94.4 \\
        && & \checkmark & 73.7 & 66.7 & 94.7 & 78.3 & 71.0 \\
        && \multirow{2}{*}{mPLUG-Owl2} & \XSolidBrush & 77.2 & 69.2 & 98.2 & 81.2 & 71.0 \\
        && & \checkmark & 89.2 & 89.2 & 89.3 & 89.2 & 50.1 \\
\midrule
\multirow{12}{*}{GQA} 
    & \multirow{4}{*}{Adversial} 
        & \multirow{2}{*}{LLaVA} & \XSolidBrush & 50.3 & 50.1 & 99.8 & 66.8 & 99.5 \\
        && & \checkmark & 54.4 & 52.5 & 93.8 & 67.3 & 89.4 \\
        && \multirow{2}{*}{mPLUG-Owl2} & \XSolidBrush & 68.4 & 63.0 & 98.2 & 75.6 & 79.8 \\
        && & \checkmark & 76.0 & 73.6 & 81.2 & 77.2 & 55.2 \\
    & \multirow{4}{*}{Popular} 
        & \multirow{2}{*}{LLaVA} & \XSolidBrush & 50.1 & 50.0 & 99.8 & 66.7 & 99.7 \\
        && & \checkmark & 58.7 & 55.1 & 94.3 & 69.5 & 85.5 \\
        && \multirow{2}{*}{mPLUG-Owl2} & \XSolidBrush & 70.6 & 63.8 & 94.9 & 76.3 & 74.4 \\

        && & \checkmark & 77.6 & 75.6 & 81.3 & 78.4 & 53.8 \\
    & \multirow{4}{*}{Random} 
        & \multirow{2}{*}{LLaVA} & \XSolidBrush & 
        55.7 & 53.0 & 99.8 & 69.2 & 94.1 \\
        && & \checkmark & 74.3 & 67.3 & 94.8 & 78.7 & 70.5 \\
        && \multirow{2}{*}{mPLUG-Owl2} & \XSolidBrush  & 82.0 & 75.2 & 95.5 & 84.1 & 63.5 \\

        && & \checkmark & 86.8 & 91.5 & 81.3 & 86.1 & 44.4 \\

\bottomrule
\end{tabular}%
}
\end{table*}

\begin{table}[!ht]
\centering
\caption{Performance on general metrics for the image captioning task, including BLEU~\citep{papineni2002bleu}, ROUGE-L~\citep{lin-2004-rouge}, CIDEr~\citep{vedantam2015cider} and SPICE~\citep{anderson2016spice} scores(\%).}
\resizebox{0.9\textwidth}{!}{%
\label{tab:ap_general_caption}
\begin{tabular}{lcccccccc}
\toprule
Model & {w/$\method$} & BLEU\_1 ($\uparrow$) &  BLEU\_2 ($\uparrow$) &  BLEU\_3 ($\uparrow$) &  BLEU\_4 ($\uparrow$) &  ROUGE\_L ($\uparrow$) &  CIDEr ($\uparrow$) & SPICE ($\uparrow$)\\
\midrule
\multirow{2}{*}{LLaVA} 
& \XSolidBrush & 14.06 & 7.12 & 3.72 & 1.90 & 22.06 & 0.08 & 16.77 \\
& \checkmark   & {18.59} & {9.96} & {5.47} & {3.04} & {26.02} & {0.21} & {20.58} \\
\midrule
\multirow{2}{*}{mPLUG-Owl2} 
& \XSolidBrush & {39.91} & {25.16} & {16.57} & {11.24} & {36.26} & {1.05} & 26.82 \\
& \checkmark   & 39.51 & 24.37 & 15.93 & 10.70 & 36.01 & 1.03 & {27.42} \\
\bottomrule
\end{tabular}%
}
\end{table}

\begin{table}[!ht]
\centering
\caption{Performance on general metrics for the LLaVA-QA90 task, including BLEU~\citep{papineni2002bleu}, ROUGE-L~\citep{lin-2004-rouge}, CIDEr~\citep{vedantam2015cider} and SPICE~\citep{anderson2016spice} scores(\%). }
\resizebox{0.9\textwidth}{!}{%
\label{tab:ap_general_qa90}
\begin{tabular}{lcccccccc}
\toprule
Model & {w/$\method$} & BLEU\_1 ($\uparrow$) &  BLEU\_2 ($\uparrow$) &  BLEU\_3 ($\uparrow$) &  BLEU\_4 ($\uparrow$) &  ROUGE\_L ($\uparrow$) &  CIDEr ($\uparrow$) & SPICE ($\uparrow$)\\
\midrule
\multirow{2}{*}{LLaVA} 
& \XSolidBrush & 21.02 & 12.91 & 8.79 & 6.41 & 32.30 & 0.93 & 31.36 \\
& \checkmark   & {23.37} & {14.39} & {9.59} & {6.83} & {33.81} & {0.99} & {31.91} \\
\midrule
\multirow{2}{*}{mPLUG-Owl2} 
& \XSolidBrush & 44.50 & 28.57 & {19.58} & {14.43} & {40.24} & {1.46} & {40.51} \\
& \checkmark   & {45.82} & {28.87} & 19.24 & 13.70 & 38.54 & 1.29 & 38.70 \\
\bottomrule
\end{tabular}%
}
\end{table}

\section{Additional Experiments}\label{app:extended_results}

\subsection{Additional Baselines}
To further contextualize the effectiveness of MARINE, we conducted additional experiments comparing our approach to a baseline that employs carefully engineered prompts designed to reduce hallucination. Specifically, we used the following prompt:

\textit{Describe the visible contents of this image in as much detail as possible without adding any information not clearly visible. Only mention objects, colors, shapes, and textures that can be directly observed in the image, avoiding assumptions about materials, functions, or contexts. If there are any uncertainties about what an object is, describe its visual characteristics (e.g., 'a circular object with a smooth surface') without inferring its purpose or identity. Avoid creative or hypothetical descriptions, and focus on observable details only.}

With two different settings:
\begin{itemize}[leftmargin=*,nosep]
\item \textit{Direct Prompting}: The original input query was replaced with the prompts as described.
\item \textit{Prompts as Additional Guidance}: We incorporated the prompt as supplemental context to guide the models in generating outputs.
\end{itemize}

\begin{table*}[!t]
    \centering
    \caption{Comparison against carefully engineered prompts.}
    \resizebox{0.85\textwidth}{!}{%
    \begin{tabular}{lccccccccc}
        \toprule
       \textbf{Method} &
        & \SetCell[c=3]{c}{{LLaVA}} & & 
        & \SetCell[c=3]{c}{{LLaVA-v1.5}} & & 
        & \SetCell[c=3]{c}{{mPLUG-Owl2}} & \\
        \cmidrule(lr){2-4} \cmidrule(lr){5-7} \cmidrule(lr){8-10}
        CHAIR & $C_s \downarrow$ & $C_i \downarrow$ & Recall $\uparrow$
        & $C_s \downarrow$ & $C_i \downarrow$ & Recall $\uparrow$
        & $C_s \downarrow$ & $C_i \downarrow$ & Recall $\uparrow$ \\
        \midrule
        Original                  & 26.6 & 10.5 & 47.4 & 8.8  & 4.6  & 41.1 & 5.0  & 3.2  & 33.2 \\
        Direct Prompting          & 27.2 & 11.0 & 46.4 & 19.6 & 8.3  & \textbf{52.3} & 9.0  & 5.1  & \textbf{42.0} \\
        Prompts as Additional Guidance & 37.4 & 10.5 & 50.4 & 12.6 & 5.9  & 44.6 & 6.6  & 3.9  & 40.4 \\
        \textbf{MARINE (ours)}    & \textbf{17.8} & \textbf{7.2} & \textbf{50.8} & \textbf{6.2} & \textbf{3.0} & 44.3 & \textbf{4.2} & \textbf{2.3} & 41.4 \\
        \bottomrule
    \end{tabular}%
    }
    \label{tab:comparison_methods}
\end{table*}

As shown in Table~\ref{tab:comparison_methods}, prompt-based guidance can improve recall for some models (e.g., LLaVA-v1.5), but does not consistently reduce hallucinations across all metrics. In fact, CHAIR scores often worsen. In contrast, MARINE achieves stronger improvements across all models.

We highlight two key differences between MARINE and prompt-based approaches:
\begin{itemize}[leftmargin=*,nosep]
\item \textit{Model Dependence}: Prompting methods rely heavily on the instruction-following capabilities of the model. While they may reduce hallucinations slightly for stronger models (e.g., LLaVA-v1.5), they can worsen performance in weaker models (e.g., LLaVA). Additionally, prompt-based approaches may require fine-tuning to be effective~\citep{deng2024enhancing}. MARINE, by contrast, improves grounding through explicit visual signals, making it effective even without model retraining.
\item \textit{Generalization and Efficiency}: Prompting methods often require task-specific tuning or dataset-aware phrasing. MARINE generalizes across tasks and models with minimal engineering and no fine-tuning, while offering more consistent hallucination reduction.
\end{itemize}

\subsection{Dynamic Guidance Strength}
\begin{table*}[!t]
    \centering
    \caption{Experiments on dynamic guidance strength based on confidence scores on CHAIR metrics.}
    \resizebox{0.55\textwidth}{!}{%
    \begin{tabular}{lcccccc}
        \toprule
        \textbf{Method} & \multicolumn{3}{c}{{LLaVA}} & \multicolumn{3}{c}{{mPLUG-Owl2}} \\
        \cmidrule(lr){2-4} \cmidrule(lr){5-7}
        CHAIR & $C_s \downarrow$ & $C_i \downarrow$ & Recall $\uparrow$ & $C_s \downarrow$ & $C_i \downarrow$ & Recall $\uparrow$ \\
        \midrule
        Fix Guidance Strength & 17.8 & 7.2 & \textbf{50.8} & \textbf{4.2} & \textbf{2.3} & \textbf{41.4} \\
        {Dynamic Guidance Strength} & \textbf{14.8} & \textbf{6.5} & 49.9 & 5.0 & 2.6 & 41.0 \\
        \bottomrule
    \end{tabular}%
    }
    \label{tab:guidance_comparison_1}
\end{table*}

\begin{table*}[!t]
    \centering
    \caption{Experiments on dynamic guidance strength based on confidence scores on POPE metrics.}
    \resizebox{0.65\textwidth}{!}{%
    \begin{tabular}{lcccccc}
        \toprule
        \textbf{Method} & \multicolumn{3}{c}{{LLaVA}} & \multicolumn{3}{c}{{mPLUG-Owl2}} \\
        \cmidrule(lr){2-4} \cmidrule(lr){5-7}
        POPE & Accuracy $\uparrow$ & F1 $\uparrow$ & Yes Ratio & Accuracy $\uparrow$ & F1 $\uparrow$ & Yes Ratio \\
        \midrule
        Fix Guidance Strength & 66.9 & 72.9 & 72.3 & 82.8 & 82.7 & 49.2 \\
        {Dynamic Guidance Strength} & \textbf{71.97} & \textbf{74.48} & \textbf{59.83} & \textbf{83.3} & \textbf{83.2} & \textbf{49.4} \\
        \bottomrule
    \end{tabular}%
    }
    \label{tab:performance_metrics}
\end{table*}

We conducted additional experiments to compare fixed and dynamic guidance strength strategies using both CHAIR and POPE metrics (Tables~\ref{tab:guidance_comparison_1} and~\ref{tab:performance_metrics}).

\begin{itemize}
    \item \textit{Fix Guidance Strength} uses a fixed guidance strength of 0.7, selected to balance hallucination reduction and instructions adherence.
    \item \textit{Dynamic Guidance Strength} adjusts the guidance strength dynamically by mapping the mean confidence score ($s$) of the image-grounding models to a range of (0.4, 0.8) using the formula 
    \begin{align*}
        \gamma' = 0.4 + \frac{(0.8 - 0.4)\cdot (s - {s_{\min}})}{s_{\max} - s_{\min}}.
    \end{align*}
\end{itemize}
A higher confidence score indicates more reliable grounding, which results in stronger guidance. Empirically, we find that dynamic guidance improves performance for weaker models such as LLaVA, which are more sensitive to noisy signals. For stronger models like mPLUG-Owl2, a fixed guidance strength is already sufficient to reduce object hallucinations effectively.

\subsection{Effect of Sampling Temperature}

In our main experiments, we use greedy decoding (temperature = 0) to ensure deterministic outputs and reproducible comparisons—consistent with our primary baseline (VCD) and common practice in hallucination benchmarks. To evaluate robustness under stochastic decoding, we also test with a temperature of 0.6 and report mean ± standard deviation in Table~\ref{tab:temp}. MARINE continues to outperform baseline generations across all hallucination metrics, demonstrating effectiveness regardless of sampling strategy.

\begin{table}[h!]
\centering
\caption{Object hallucination metrics under temperature = 0.6 sampling.}
\label{tab:temp}
\resizebox{0.6\linewidth}{!}{
\begin{tabular}{lcccccc}
\toprule
\textbf{Method} 
& \multicolumn{3}{c}{{LLaVA}} 
& \multicolumn{3}{c}{{mPLUG-Owl2}} \\
\cmidrule(lr){2-4} \cmidrule(lr){5-7}
CHAIR & $C_s\downarrow$ & $C_i\downarrow$ & Recall$\uparrow$ 
& $C_s\downarrow$ & $C_i\downarrow$ & Recall$\uparrow$ \\
\midrule
Greedy 
& 26.1$_{\pm1.6}$ & 10.8$_{\pm0.5}$ & 46.0$_{\pm0.8}$ 
& 4.9$_{\pm0.6}$ & 2.8$_{\pm0.3}$ & 37.7$_{\pm0.6}$ \\
MARINE (ours) 
& \textbf{19.3}$_{\pm0.8}$ & \textbf{7.6}$_{\pm0.1}$ & \textbf{50.6}$_{\pm0.2}$ 
& \textbf{4.5}$_{\pm0.6}$ & \textbf{2.4}$_{\pm0.2}$ & \textbf{41.1}$_{\pm0.4}$ \\
\bottomrule
\end{tabular}
}
\end{table}

\subsection{Memory Analysis}

We evaluated the peak GPU memory usage during inference on 500 image captioning examples using the LLaVA model, with a batch size of 16 and a maximum generation length of 64 tokens. Results are reported in Table~\ref{tab:memory_app}.
Although MARINE introduces additional vision models, the overall memory footprint increases by only approximately 30\% during inference—significantly less than doubling. This is because the object detection models are relatively lightweight compared to the large LLM backbone.

\begin{table*}[!t]
\centering
\caption{Peak GPU Memory Usage during Inference (GB) of $\method$ compared to greedy decoding and VCD.}
\label{tab:memory_app}
\resizebox{0.55\linewidth}{!}{
\begin{tabular}{lccc}
    \toprule
    \textbf{Metric} & Greedy & VCD & $\method$ (Ours) \\
    \midrule
    Peak GPU Memory Usage & 23.53 & 20.73 ($\times$0.88) & 30.78 ($\times$1.30) \\
    \bottomrule
\end{tabular}
}
\end{table*}

\subsection{Further Study on Guidance Strength}
\label{app:ab_guidance_strength}

\begin{figure}[ht!]
    \centering
    \includegraphics[width=0.6\textwidth]{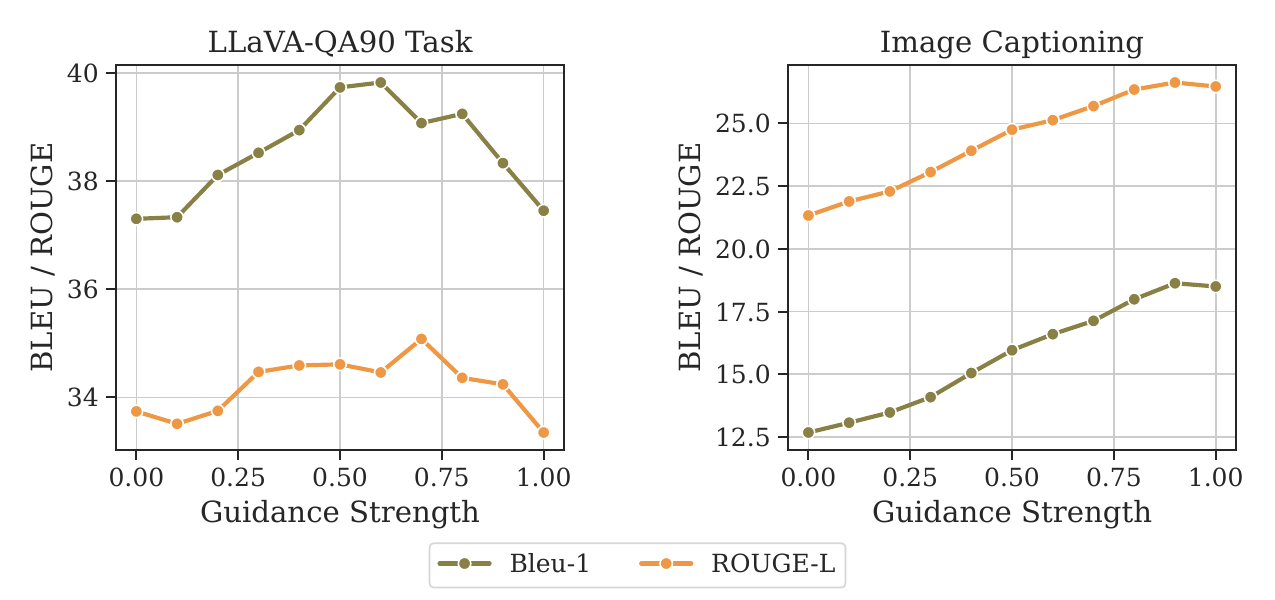}
    \caption{The impact of guidance strength on the output text quality.}
    \label{fig:ap_ablation_quality}
\end{figure}

Figure~\ref{fig:ap_ablation_quality} shows how varying the guidance strength $\gamma$ affects the quality of LLaVA’s output on the LLaVA-QA90 and image captioning tasks (max generation length = 256). We observe that setting $\gamma = 1$ does not yield the best image captioning performance. In the LLaVA-QA90 task, guidance strengths in the range of 0.5 to 0.7 lead to higher output quality. This observation is consistent with prior findings in classifier-free guidance literature: overly strong guidance can dominate the generation process and reduce fluency or instruction adherence.

\begin{figure}[ht!]
    \centering
    \includegraphics[width=\columnwidth]{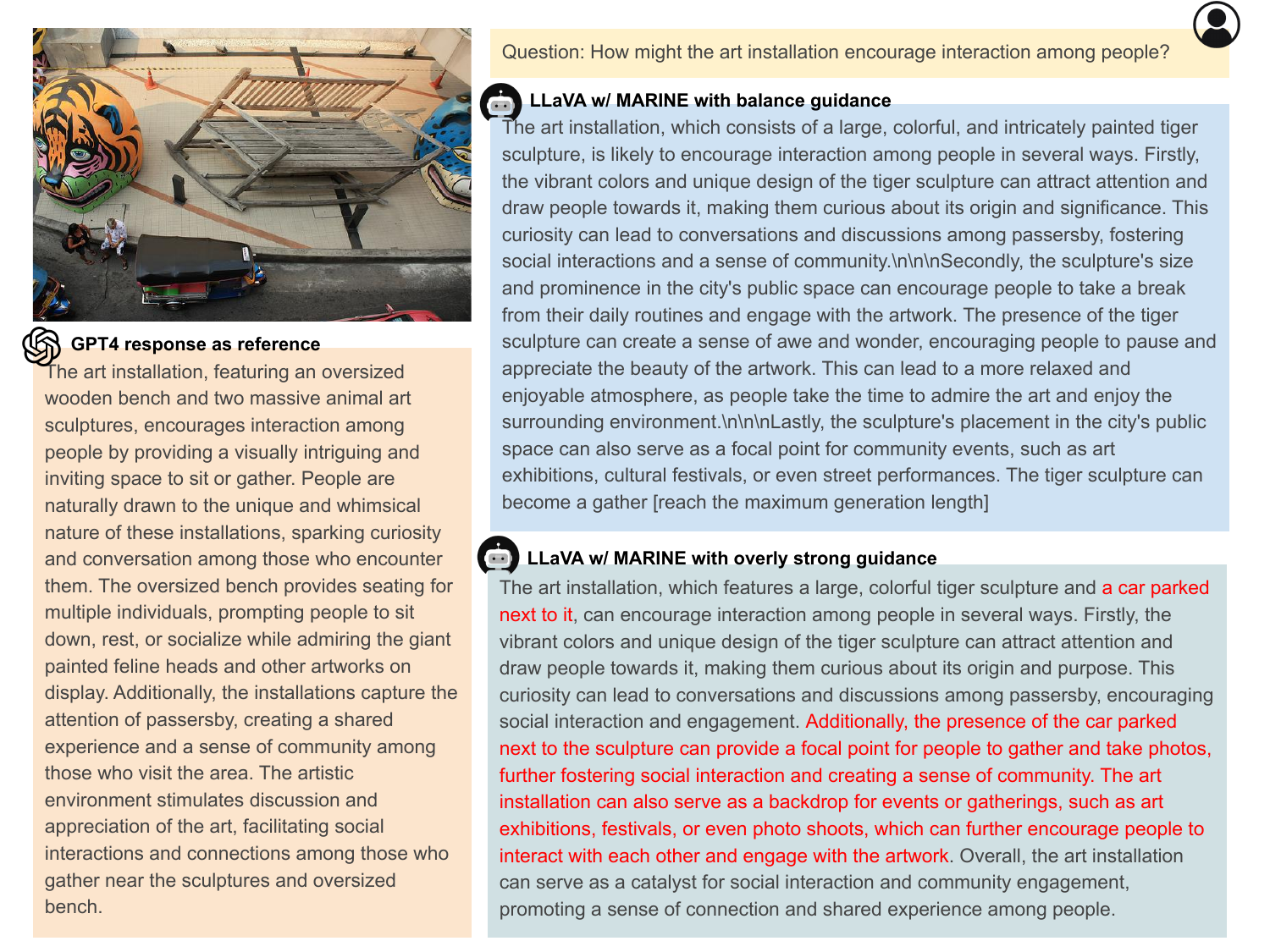}
    \caption{This case highlights that overly strong guidance can induce the model to prioritize providing exhaustive visual details from the image, even when such details are irrelevant to the specific instruction (e.g., ``a car parked next to it''). In contrast, balanced guidance enables the model to maintain better adherence to the instruction while still utilizing the visual information effectively.}
    \label{fig:ap_example_gamma}
\end{figure}

To further validate these results, we use GPT-4V as an automatic judge to score outputs (on a 10-point scale) for accuracy and detail. The results, summarized in Table~\ref{tab:gpt4v-aided-app}, show that balancing the original LVLM branch leads to improved generation quality.
Finally, Figure~\ref{fig:ap_example_gamma} provides qualitative examples showing how excessive guidance can reduce instruction alignment, often introducing unnecessary visual details into the response.

\begin{table*}[!t]
\centering
\caption{Results of GPT-4V-aided evaluation. The accuracy and detailedness metrics are on a scale of 10, and a higher score indicates better performance. The symbols \(\times\) and \(\checkmark\) indicate performance metrics without and with our method, respectively.}
\label{tab:gpt4v-aided-app}
\resizebox{0.4\textwidth}{!}{
\begin{tabular}{llcc}
\toprule
\textbf{Task} & \textbf{Metric $\uparrow$} & \textbf{\XSolidBrush~($\gamma=1$)} & \textbf{\checkmark~($\gamma=0.7$)} \\
\midrule
\multirow{2}{*}{LLaVA-QA90} 
& Accuracy & 5.52 & \textbf{5.79} \\
& Detailedness & 4.58 & \textbf{4.77} \\
\midrule
\multirow{2}{*}{Image Captioning} 
& Accuracy & 6.06 & \textbf{6.22} \\
& Detailedness & 5.00 & \textbf{5.24} \\
\bottomrule
\end{tabular}
}
\end{table*}

\newpage
\section{Further Analysis}\label{app:exp_results}

\subsection{Limitations of Hallucination Evaluation}

While CHAIR and POPE are widely adopted for evaluating object hallucinations in vision-language models, both have inherent limitations. CHAIR depends on a fixed object vocabulary and synonym list, which may miss rare or fine-grained concepts. POPE relies on the quality of segmentation tools to define ground-truth objects, introducing variability across settings.

To address these limitations, we incorporate \textit{ALOHa} (Automatic Localized Hallucination)~\citep{petryk-etal-2024-aloha}, a reference-based metric that evaluates hallucination at both the object level ($ALOHa_0$) and the caption level ($ALOHa$). We follow the standard ALOHa setup using MSCOCO ground-truth captions and enable reference object detection for more precise and generalizable assessment. As shown in Table~\ref{tab:aloha_results}, MARINE consistently outperforms greedy decoding across all models and both ALOHa metrics.

\begin{table}[h!]
\centering
\caption{ALOHa hallucination scores (all values are in \%). MARINE improves over greedy decoding across models and metrics.}
\label{tab:aloha_results}
\resizebox{0.7\linewidth}{!}{
\begin{tabular}{lcccccc}
\toprule
\textbf{Method} 
& \multicolumn{2}{c}{{LLaVA}} 
& \multicolumn{2}{c}{{LLaVA-v1.5}} 
& \multicolumn{2}{c}{{mPLUG-Owl2}} \\
\cmidrule(lr){2-3} \cmidrule(lr){4-5} \cmidrule(lr){6-7}
& $ALOHa\uparrow$ & $ALOHa_0\uparrow$ 
& $ALOHa\uparrow$ & $ALOHa_0\uparrow$ 
& $ALOHa\uparrow$ & $ALOHa_0\uparrow$ \\
\midrule
Greedy 
& 40.1 & 70.1 
& 61.9 & 83.1 
& 70.2 & 87.0 \\
MARINE 
& \textbf{48.7} 
& \textbf{76.1}
& \textbf{66.7}
& \textbf{85.6}
& \textbf{72.9}
& \textbf{88.2} \\
\bottomrule
\end{tabular}
}
\end{table}

\subsection{Additional Related Work}
\label{app:related_work}

Several recent works aim to improve grounding or reduce hallucination in vision-language models. BRAVE~\citep{kar2024bravebroadeningvisualencoding} enhances faithfulness by combining diverse visual sources, similar in spirit to MARINE, but introduces additional trainable modules. MARINE achieves comparable performance with a training-free, modular design.

Other approaches focus on evaluation~\citep{hu2023tifaaccurateinterpretabletexttoimage, cho2024davidsonianscenegraphimproving, lin2024evaluatingtexttovisualgenerationimagetotext} or feature-level interventions~\citep{yang2025nullumitigatingobjecthallucinations, liu2024reducinghallucinationsvisionlanguagemodels} to steer models away from hallucinations. \citet{liu2024paying} address \textit{text inertia}, where models generate similar outputs regardless of image content. \citet{wan2024contrastive} introduce sub-image contrastive alignment, and \citet{zhang2024prompt} control generation by adjusting visual attention weights.

These methods highlight complementary strategies to MARINE’s structured, object-level guidance for reducing hallucination.

\subsection{Effect of $\method$ on logit distribution.}
\begin{figure}[ht!]
\centering 
\subfigure[An example of image description where the original LLaVA outputs a hallucinated object, ``fork''.]{\label{fig:llava_MARINE_sample}\includegraphics[width=0.48\textwidth]{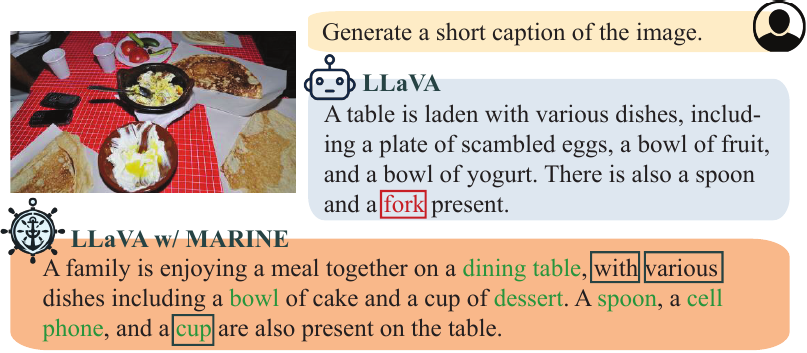}}
\subfigure[The probability distributions at the token of the hallucinated word in the original, control, and $\method$ outputs. $\method$ effectively decreases the the probability of ``fork''.]{\label{fig:probs_hallu}\includegraphics[width=0.48\textwidth]{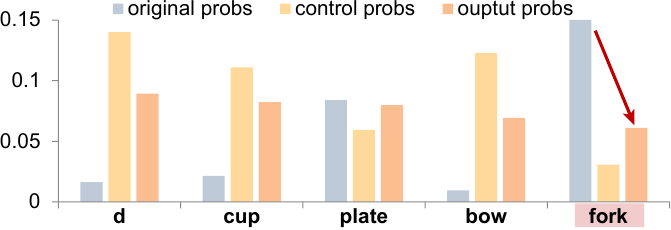}}
\subfigure[Probabilities of non-hallucinated words remain the same, highlighting $\method$'s ability to preserve normal outputs.]{\label{fig:probs_non_hallu}\includegraphics[width=0.5\textwidth]{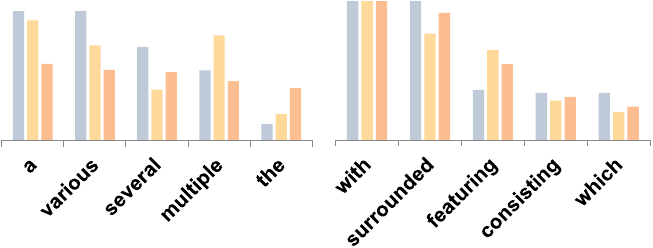}}
\caption{This sample shows how $\method$ controls logit distributions to mitigate hallucinations like ``fork'' while preserving the probabilities of ``with'', ``various'' during generation.}
\label{fig:ablation_logits_alternation}
\end{figure}

In Figure~\ref{fig:ablation_logits_alternation}, we illustrate a specific example that shows how $\method$ influences the logit distribution of LVLMs during text generation. 
Specifically, $\method$ is observed to selectively target the potential hallucinated tokens, reducing their original probabilities to mitigate the risk of hallucination in the generated text. 
For instance, in the provided example, the probability of ``fork'' is significantly lowered with $\method$, which would have originally resulted in a hallucinated object. 
Conversely, standard language elements such as ``various'', an adjective describing the overall image context, and ``with'', a crucial preposition, maintain their original probabilities. 
This selective nature of modulation by $\method$ ensures coherent and contextually relevant text generation that adheres to the instruction while effectively reducing hallucinations. 

\subsection{Discussion on fine-tuning methods.} 
\begin{figure}[ht!]
    \centering
    \includegraphics[width=0.65\columnwidth]{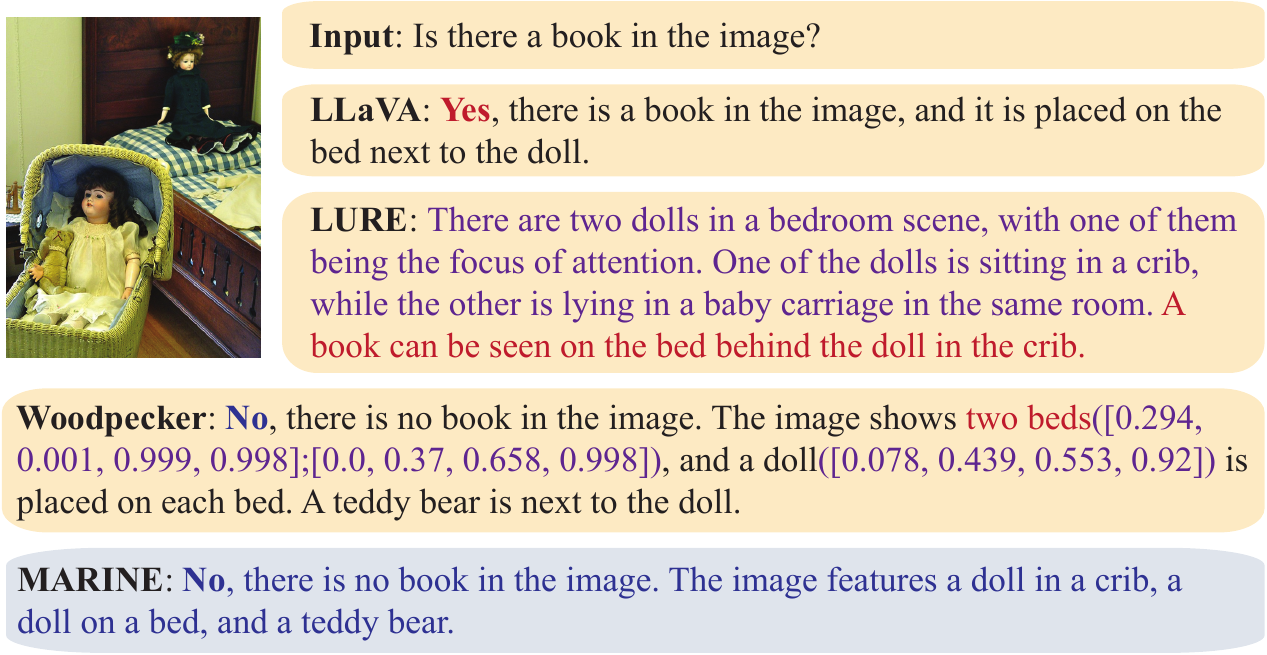}
    \caption{Example responses to an image-question pair. The LURE-corrected output deviates from the original question, offering irrelevant descriptions without directly addressing the query. Woodpecker hallucinates the existence of two beds while there is only one bed in the figure. In contrast, $\method$ maintains the original answer's style and adheres to the user's instruction while eliminating hallucination.}
    \label{fig:example}
\end{figure}
The examples shown in Figure~\ref{fig:example} illustrate that LURE, at times, fails to adhere to the given instructions when correcting LVLM generations. 
Despite receiving concise image descriptions generated based on instructions for short responses, LURE predominantly overwrites them with excessively long responses that contain information irrelevant to the instruction.  
Furthermore, LURE fails to adequately address the binary question format of POPE, as LURE fixates on extended descriptions without responding with ``yes'' or ``no'', making its evaluation using POPE impractical. 
This issue can be prevalent in small-scale fine-tuning methods, where the limited variety of the specifically tailored fine-tuning dataset harms the model's performance on other tasks. 
In contrast, the training-free approach of $\method$ demonstrates effective mitigation of hallucinations across a variety of question formats. 

\subsection{Extended Analysis in Ablation Study}
\begin{figure}[ht!]
\centering   
\subfigure[CHAIR$_S$]{\label{fig:ab2_llava_chair_s}\includegraphics[width=0.20\textwidth]{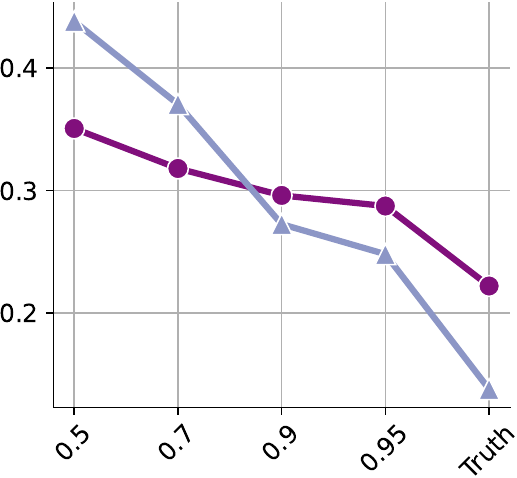}}
\subfigure[CHAIR$_I$]{\label{fig:ab2_llava_chair_i}\includegraphics[width=0.20\textwidth]{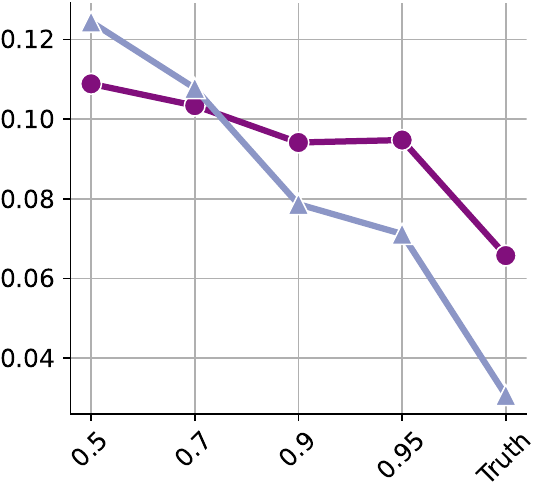}}
\subfigure[Recall]{\label{fig:ab2_llava_recall}\includegraphics[width=0.20\textwidth]{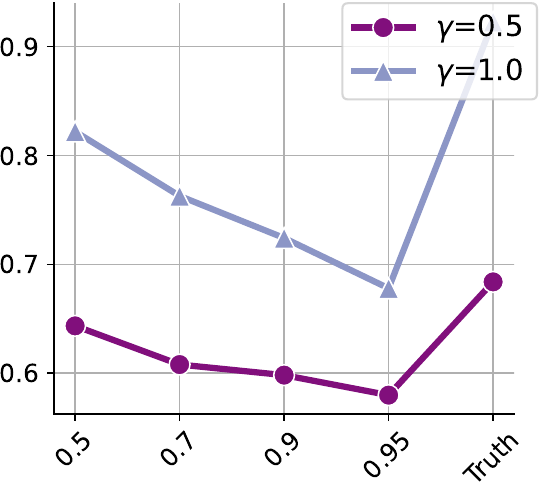}}
\caption{LLaVA's performance on CHAIR according to different score threshold of object grounding features in $\method$. We consider four confidence thresholds ($0.5$, $0.7$, $0.9$, and $0.95$) for DETR to vary the score threshold.}
\label{fig:ablation_threshold}
\end{figure}

\begin{figure}[ht!]
\centering   
\subfigure[CHAIR$_S$]{\label{fig:ap2_mplug_chair_s}\includegraphics[width=0.20\textwidth]{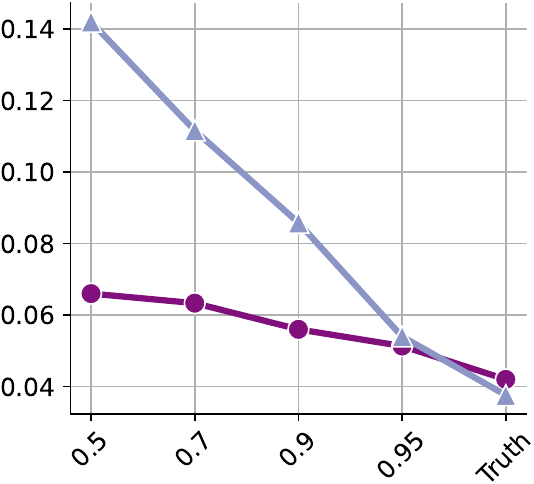}}
\subfigure[CHAIR$_I$]{\label{fig:ap2_mplug_chair_i}\includegraphics[width=0.20\textwidth]{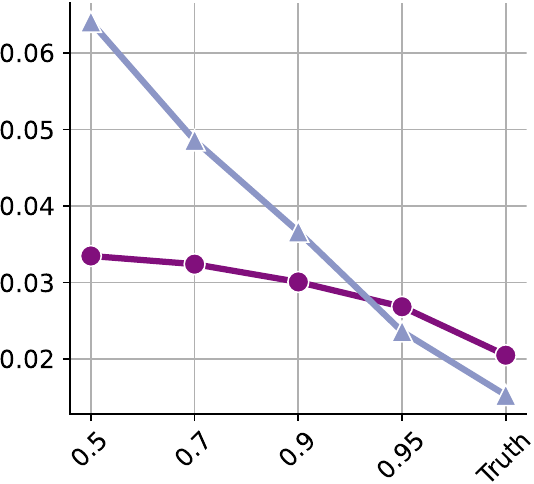}}
\subfigure[Recall]{\label{fig:ap2_mplug_recall}\includegraphics[width=0.20\textwidth]{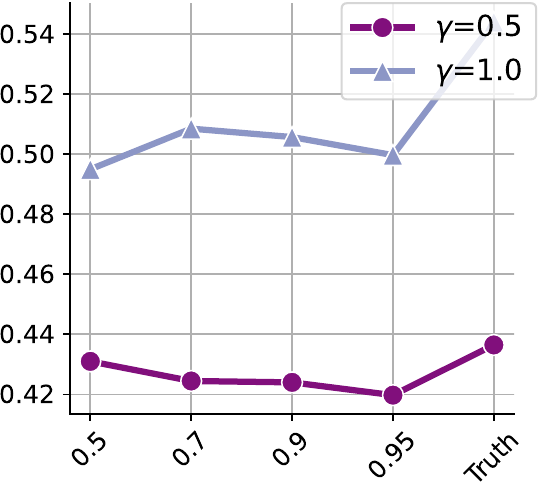}}
\caption{mPLUG-Owl2's performance on CHAIR according to different score threshold of object grounding features in $\method$. We consider four confidence thresholds ($0.5$, $0.7$, $0.9$, and $0.95$) for DETR to vary the score threshold, with $\method$-Truth serving as an ideal reference.}
\label{fig:ap2_owl2}
\end{figure}

Additional experimental results explore the score threshold of object grounding features, which are examined across LLaVA, and mPLUG-Owl2, with findings presented in Figures~\ref{fig:ablation_threshold}, and~\ref{fig:ap2_owl2}.

This variation is achieved by implementing four confidence thresholds ($0.5$, $0.7$, $0.9$, and $0.95$) in the DETR model predictions (with $\method$-Truth serving as an ideal reference), where higher thresholds correspond to lesser, yet higher-quality, visual information. 
Our findings highlight two significant insights. 
Firstly, an increase in the quality of visual information correlates with a noticeable decrease in hallucinations produced by the LVLMs. 
A lower threshold, which allows for more visual information but also includes noisier content, could potentially result in an increased occurrence of hallucinations. 
Furthermore, lower-quality visual information is associated with enhanced Recall. 
This suggests that LVLMs under guidance, despite the presence of noisy visual inputs, tend to focus more on the visual details (i.e., objects), resulting in more elaborate descriptions.

\subsection{More Case Studies}

\begin{figure}[t!]
    \centering
    \includegraphics[width=\linewidth]{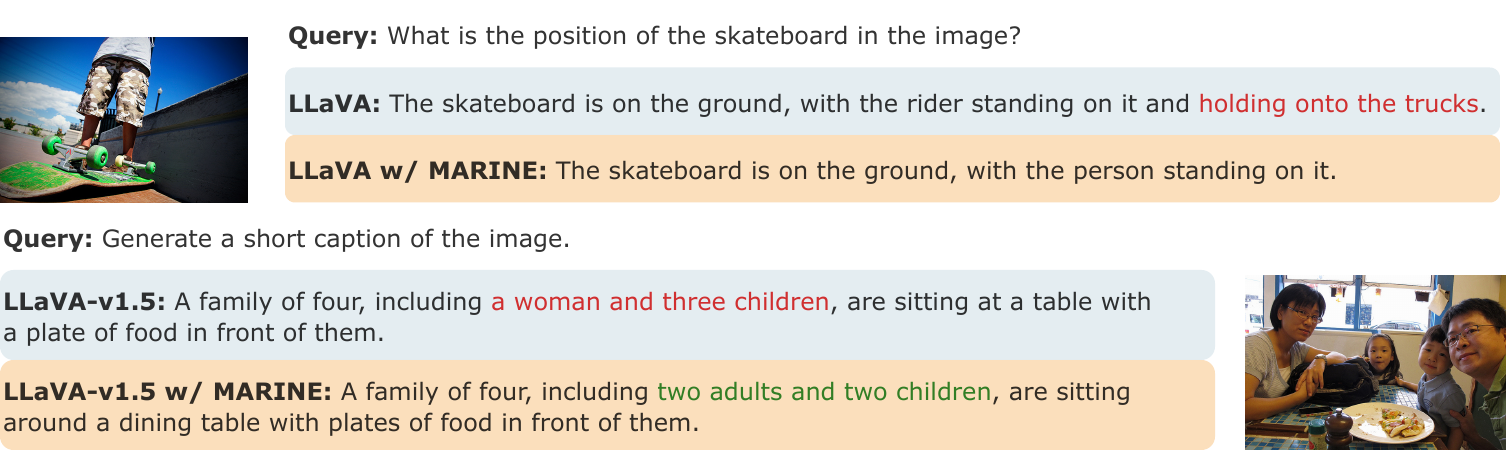}
    \caption{Hallucination mitigation examples by our proposed $\method$ across multiple tasks: LLaVA-QA90 and image captioning. Hallucinated objects generated by the LVLM are highlighted in red.}
    \label{fig:examples_appendix}
\end{figure}

\begin{figure}[ht!]
    \centering
\includegraphics[width=\textwidth]{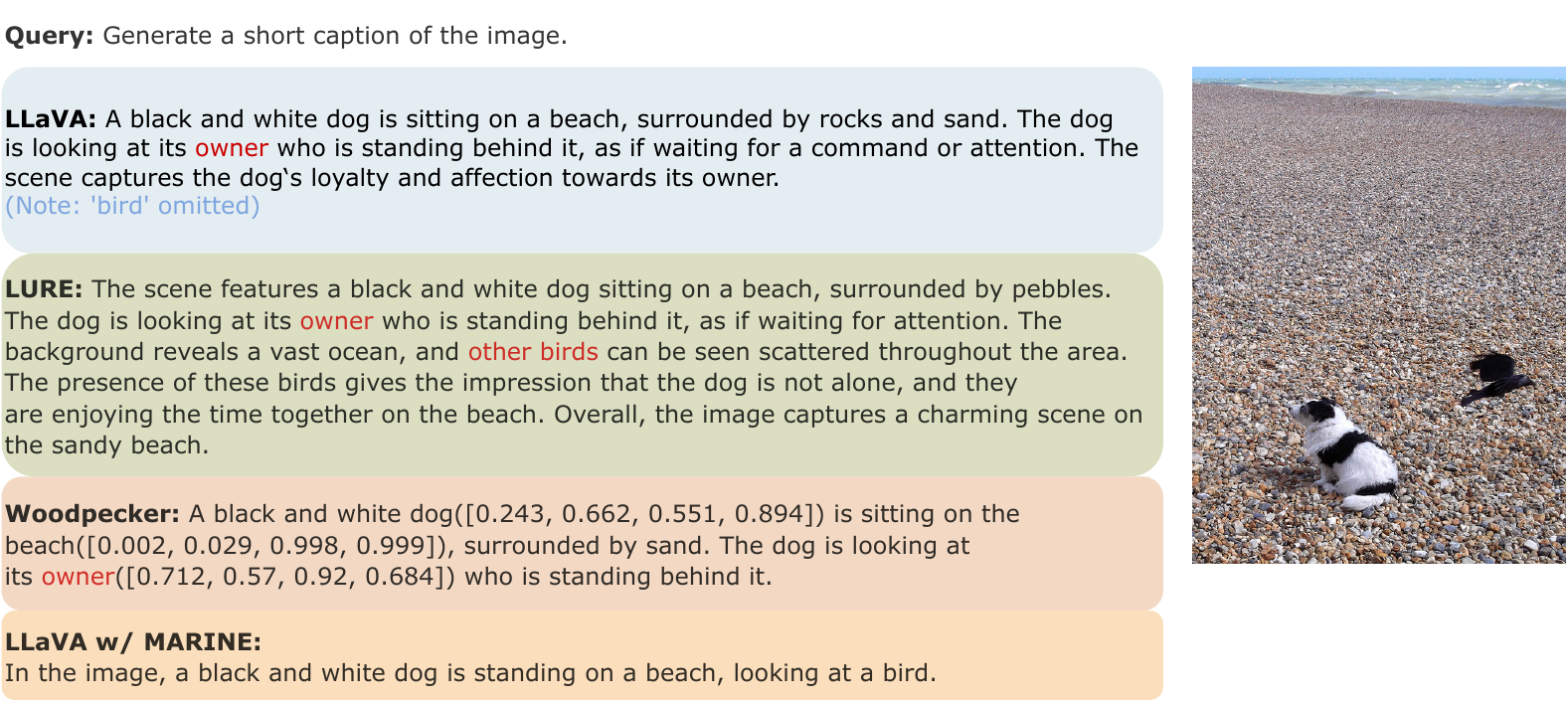}
    \caption{A comparison of responses from baseline models and our $\method$ in an image description task. It illustrates $\method$'s superior ability to reduce hallucinations, in contrast to LURE and Woodpecker, which fail to effectively address hallucinations and sometimes even increase hallucinated content. This example highlights the strengths of our correct-during-generation framework over post-correction approaches, showcasing its efficiency, preservation of original style, and enhanced adherence to instructions.}
    \label{fig:sp_chair1}
\end{figure}

\begin{figure}[ht!]
    \centering
    \includegraphics[width=\textwidth]{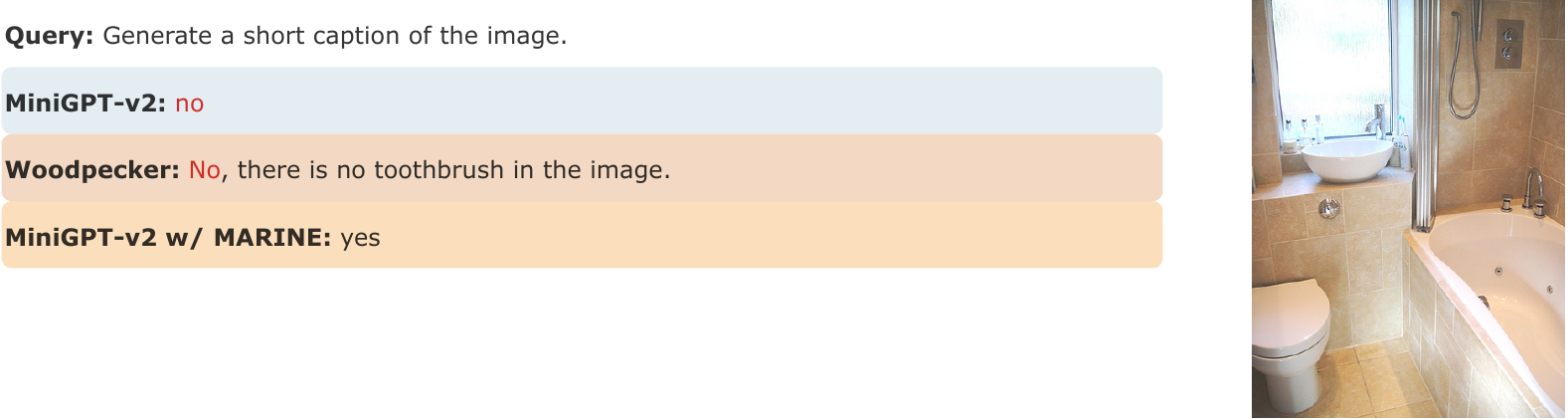}
    \caption{A comparison of responses from baseline models and our $\method$ in POPE ``yes-or-no'' task. 
    MiniGPT-v2 provides a concise response without referencing any objects. Under these circumstances, Woodpecker is unable to perform corrections via GPT-3.5 due to missing visual details. $\method$, however, successfully corrects the response while retaining MiniGPT-v2's style.}
    \label{fig:sp_pope1}
\end{figure}

In Figures~\ref{fig:examples_appendix}, ~\ref{fig:sp_chair1} and~\ref{fig:sp_pope1}, we present examples of the outputs from LURE~\citep{zhou2023analyzing}, Woodpecker~\citep{yin2023woodpecker} and $\method$ on different tasks further validate our arguments in the paper. 

\end{document}